\documentclass{egpubl}
\usepackage{eg2024}
 
\ConferencePaper        %

\CGFccby

\usepackage[T1]{fontenc}
\usepackage{dfadobe}  

\usepackage{cite}  %
\BibtexOrBiblatex
\electronicVersion
\PrintedOrElectronic

\ifpdf \usepackage[pdftex]{graphicx} \pdfcompresslevel=9
\else \usepackage[dvips]{graphicx} \fi

\usepackage{egweblnk} 

\usepackage{times}
\usepackage{epsfig}
\usepackage{graphicx}
\usepackage{amsmath}
\usepackage{xcolor}
\usepackage{tabularx}
\usepackage{capt-of}
\usepackage{enumitem} %
\usepackage{booktabs} %
\usepackage{soul}
\usepackage{pifont} %
\usepackage{color}
\usepackage{colortbl}
\usepackage{multirow}
\usepackage[normalem]{ulem}

\newcommand{\cmark}{\text{\ding{51}}} %
\newcommand{\xmark}{\text{\ding{55}}} %

\usepackage[ruled]{algorithm2e} %

\SetAlFnt{\small}
\SetAlCapFnt{\small}
\SetAlCapNameFnt{\small}
\SetAlCapHSkip{0pt}

\newcommand{\ourmethod}{HaLo-NeRF}
\newcommand{\ourdataset}{HolyScenes}
\newcommand{\ourclip}{CLIP\textsubscript{FT}}
\newcommand{\ourclipseg}{CLIPSeg\textsubscript{FT}}
\newcommand{\ourlerf}{LERF\textsubscript{FT}}
\long\def\ignorethis#1{}

\makeatletter
\DeclareRobustCommand\onedot{\futurelet\@let@token\@onedot}
\def\@onedot{\ifx\@let@token.\else.\null\fi\xspace}

\def\etal{\emph{et al}\onedot}

\usepackage{color}
\definecolor{gray}{rgb}{0.6,0.6,0.6}
\definecolor{red}{rgb}{1,0,0}
\definecolor{green}{rgb}{0,1,0}
\definecolor{blue}{rgb}{0,0,1}
\definecolor{dark-green}{rgb}{0,0.4,0}
\definecolor{orange}{rgb}{1,0.55,0}
\definecolor{white}{rgb}{1,1,1}
\definecolor{black}{rgb}{0,0,0}
\definecolor{dark-brown}{rgb}{0.2,0.1,0}
\definecolor{light-blue}{rgb}{0.4,0.6,0.99}
\definecolor{dark-red}{rgb}{0.6,0,0}
\definecolor{light-red}{rgb}{1,0.2,0.6}
\definecolor{pink}{rgb}{1,0.2,0.6}
\definecolor{dark-pink}{rgb}{0.6,0,0.3}

\newcommand{\whitetxt}[1]{{\color{white}#1}\normalfont}

\newcommand{\B}{$\left<\textsc{building}\right>$\xspace}

\newbox\jsavebox
\newcommand{\jsubfig}[2]{%
	\sbox\jsavebox{#1}%
	\parbox[t]{\wd\jsavebox}{\centering\usebox\jsavebox\\#2}%
	}

\newcommand{\sayitt}[3]{{\small\protect\colorlet{col}{#2}\color{col}{} #3}}

    \newcommand{\archway}[1]{\sayitt{}{cyan}{#1}}
    \newcommand{\facade}[1]{\sayitt{}{orange}{#1}}
    \newcommand{\sundial}[1]{\sayitt{}{teal}{#1}}

\newcommand{\new}[1]{{\textcolor{black}{#1}}}

\newcolumntype{C}{>{\centering\arraybackslash}X}

\begin{document}

\title{\new{\ourmethod{}: Learning Geometry-Guided Semantics  \\ \, for Exploring Unconstrained Photo Collections}  } %

\author[C. Dudai et al.]
{\parbox{\textwidth}{\centering Chen Dudai*$^{1}$,\: Morris Alper*$^{1}$,\: Hana Bezalel$^{1}$,\: Rana Hanocka$^{2}$,\: Itai Lang$^{2}$, \: Hadar Averbuch-Elor$^{1}$ 
        }
\\
{\parbox{\textwidth}{\centering
$^1$Tel Aviv University \quad
        $^2$University of Chicago 
       }
}
}

\teaser{
\centering
\includegraphics[width=\linewidth]{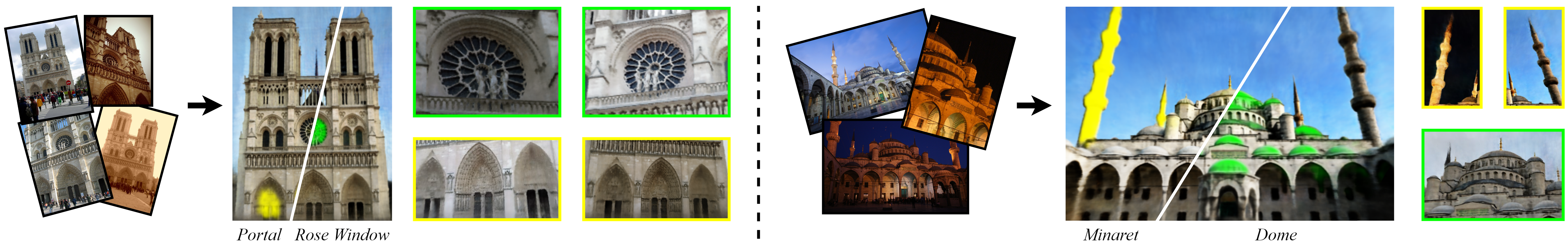}
\caption{Given a collection of images in-the-wild depicting a large-scale scene, such as the Notre-Dame Cathedral or the Blue Mosque above, we learn a semantic localization field for each textual description (shown with green and yellow overlay). Our approach enables generating novel views with controlled appearances of these semantic regions of interest (as shown in the boxes of corresponding colors).}
\label{fig:teaser}
}

\maketitle

\begin{abstract}
\new{Internet image collections containing photos captured by crowds of photographers show promise for enabling digital exploration of large-scale tourist landmarks. However, prior works focus primarily on geometric reconstruction and visualization, neglecting the key role of language in providing a semantic interface for navigation and fine-grained understanding. 
In more constrained 3D domains, recent methods have leveraged modern vision-and-language models as a strong prior of 2D visual semantics. } 
\new{While these models display an excellent understanding of broad visual semantics, they struggle with unconstrained photo collections depicting such tourist landmarks, as they lack expert knowledge of the architectural domain and fail to exploit the geometric consistency of images capturing multiple views of such scenes.}
\new{In this work, we present a localization system
that connects neural representations of scenes depicting large-scale landmarks with text describing a semantic region within the scene, by harnessing the power of SOTA vision-and-language models with adaptations for understanding landmark scene semantics.}
To bolster such models with fine-grained knowledge, we leverage large-scale Internet data containing images of similar landmarks along with weakly-related textual information. Our approach is built upon the premise that images physically grounded in space can provide a powerful supervision signal for localizing new concepts, whose semantics may be unlocked from Internet textual metadata with large language models. We use correspondences between views of scenes to bootstrap spatial understanding of these semantics, providing guidance for 3D-compatible segmentation that ultimately lifts to a volumetric scene representation.
To evaluate our method, we present a new benchmark dataset containing large-scale scenes with ground-truth segmentations for multiple semantic concepts.
Our results show that \ourmethod{} can accurately localize a variety of semantic concepts related to architectural landmarks, surpassing the results of other 3D models as well as strong 2D segmentation baselines.
Our code and data are publicly available at \href{https://tau-vailab.github.io/HaLo-NeRF}{\url{https://tau-vailab.github.io/HaLo-NeRF/}}.



\end{abstract}

\section{Introduction} \label{sec:introduction}

Our world is filled with incredible buildings and monuments that contain a rich variety of architectural details. Such intricately-designed human structures have attracted the interest of tourists and scholars alike. Consider, for instance, the Notre-Dame Cathedral pictured above. 
\def\thefootnote{*}\footnotetext{Denotes equal contribution}
This monument is visited annually by over 10 million people from all around the world. While Notre-Dame's facade is impressive at a glance, its complex architecture and history contain details which the untrained eye may miss. Its structure includes features such as portals, towers, and columns, as well as more esoteric items like \emph{rose window} and \emph{tympanum}. Tourists often avail themselves of guidebooks or knowledgeable tour guides in order to fully appreciate the grand architecture and history of such landmarks. But what if it were possible to explore and understand such sites without needing to hire a tour guide or even to physically travel to the location?

The emergence of neural radiance fields presents new possibilities for creating and exploring virtual worlds that contain such large-scale monuments, without the (potential burden) of traveling.
Prior work, including NeRF-W~\cite{martin2021nerf} and Ha-NeRF~\cite{chen2022hallucinated}, has demonstrated that photo-realistic images with independent control of viewpoint and illumination can be readily rendered from unstructured imagery for sites such as the Notre-Dame Cathedral. 
However, these neural techniques lack the high-level semantics embodied within the scene---such semantic understanding is crucial for exploration of a new place, similarly to the travelling tourist.

\new{Recent progress in language-driven 3D scene understanding has leveraged strong two-dimensional priors provided by modern vision-and-language (V\&L) representations~\cite{huang2022multi,chen2022ham,chen2022language,kobayashi2022decomposing,kerr2023lerf}. However, while existing pretrained vision-and-language models (VLMs) show broad semantic understanding, architectural images use a specialized vocabulary of terms (such as the \emph{minaret} and \emph{rose window} depicted in Figure \ref{fig:teaser}) that is not well encapsulated by these models out of the box.}
Therefore, we propose an approach for performing semantic adaptation of VLMs by leveraging Internet collections of landmark images and textual metadata. Inter-view coverage of a scene provides richer information than collections of unrelated imagery, as observed in prior work utilizing collections capturing physically grounded in-the-wild images~\cite{wang2020learning,iqbal2020weakly,wu2021towers}. Our key insight is that modern foundation models allow for extracting a powerful supervision signal from \emph{multi-modal} data depicting large-scale tourist scenes.

To unlock the relevant semantic categories from noisy Internet textual metadata accompanying images, we leverage the rich knowledge of large language models (LLMs). We then localize this \emph{image-level} semantic understanding to \emph{pixel-level} probabilities by leveraging the 3D-consistent nature of our image data. In particular, by bootstrapping with inter-view image correspondences, we fine-tune an image segmentation model to both learn these specific concepts and to localize them reliably within scenes, providing a 3D-compatible segmentation.

\new{We demonstrate the applicability of our approach for connecting low-level neural representations depicting such real-world tourist landmarks with higher-level semantic understanding. Specifically, we present a text-driven localization technique that is supervised on our image segmentation maps, which augments the recently proposed Ha-NeRF neural representation~\cite{chen2022hallucinated} with a localization head that predicts volumetric probabilities for a target text prompt. 
By presenting the user with a visual \emph{halo} marking the region of interest, our approach provides an intuitive interface for interacting with virtual 3D environments depicting architectural landmarks. \ourmethod{} (Ha-NeRF + \textbf{Lo}calization halo) allows the user to ``zoom in'' to the region containing the text prompt and view it from various viewpoints and across different appearances, yielding a substantially more engaging experience compared to today's common practice of browsing thumbnails returned by an image search.}

To quantitatively evaluate our method, we introduce \emph{\ourdataset{}}, a new benchmark dataset composed of six places of worship annotated with ground-truth segmentations for multiple semantic concepts. We evaluate our approach qualitatively and quantitatively, including comparisons to existing 2D and 3D techniques. Our results show that \ourmethod{} allows for localizing a wide array of elements belonging to structures reconstructed in the wild, capturing the unique semantics of our use case and significantly surpassing the performance of alternative methods.

Explicitly stated, our key contributions are:
\begin{itemize}[nolistsep]
    \item A novel approach for performing semantic adaptation of VLMs which leverages inter-view coverage of scenes in multiple modalities (namely textual metadata and geometric correspondences between views) to bootstrap spatial understanding of domain-specific semantics;
    \item A system enabling text-driven 3D localization of large-scale scenes captured in-the-wild;
    \item Results over diverse scenes and semantic regions, and a benchmark dataset for rigorously  evaluating the performance of our system as well as facilitating future work linking Internet collections with a semantic understanding.
\end{itemize}

\section{Related Work} \label{sec:related_work}

\begin{figure*}[!t]
\centering
    \jsubfig{\includegraphics[height=2.5cm]{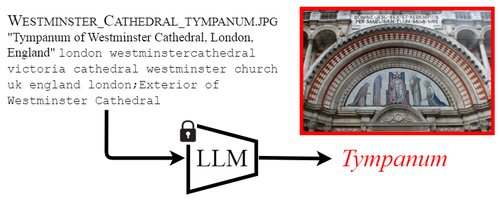}}
    {(a) LLM-Based Semantic Concept Distillation}
    \hfill
    \jsubfig{\includegraphics[height=2.5cm]{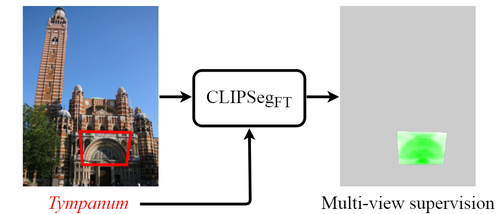}}
    {(b) Semantic Adaptation of V\&L Models}
    \hfill
    \jsubfig{\includegraphics[height=2.5cm]{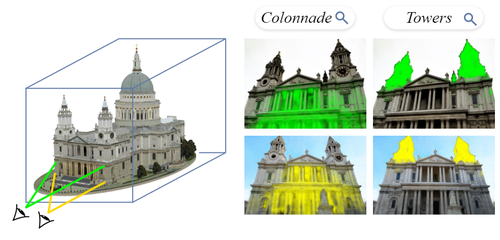}}
    {(c) Text-Driven 3D Localization}
\caption{\textbf{System overview of our approach.} (a) We extract semantic pseudo-labels from noisy Internet image metadata using a large language model (LLM). (b) We use these pseudo-labels and correspondences between scene views to learn image-level and pixel-level semantics. In particular, we fine-tune an image segmentation model (\ourclipseg{}) using multi-view supervision---where zoomed-in views and their associated pseudo-labels (such as image on the left associated with the term ``tympanum'') provide a supervision signal for zoomed-out views. (c) We then lift this semantic understanding to learn volumetric probabilities over new, unseen landmarks (such as the St. Paul's Cathedral depicted on the right), allowing for rendering views of the segmented scene with controlled viewpoints and illumination settings. See below for the definitions of the concepts shown\setcounter{footnote}{0}\protect\footnotemark{}.
}
\label{fig:system}
\end{figure*}

\noindent \textbf{Text-guided semantic segmentation.} The emergence of powerful large-scale vision-language models~\cite{jia2021scaling, radford2021learning} has propelled a surge of interest in pixel-level semantic segmentation using text prompts~\cite{xu2021simple, li2022language, luddecke2022image, ding2022decoupling, xu2022groupvit, ghiasi2022scaling, zhou2022maskclip, liang2023open}. A number these works leverage the rich semantic understanding of CLIP~\cite{radford2021learning}, stemming from large-scale contrastive training on text-image pairs.

LSeg~\cite{li2022language} trains an image encoder to align a dense pixel representation with CLIP's embedding for the text description of the corresponding semantic class. OpenSeg~\cite{ghiasi2022scaling} optimizes a class-agnostic region segmentation module to matched extracted words from image captions. 
CLIPSeg~\cite{luddecke2022image} leverages the activations of CLIP's dual encoders, training a decoder to convert them into a binary segmentation mask.
CLIP's zero-shot understanding on the image level has also been leveraged for localization by Decatur \etal~\cite{decatur20223d}, who lift CLIP-guided segmentation in 2D views to open-vocabulary localization over 3D meshes.

These methods aim for general open-vocabulary image segmentation and can achieve impressive performance over a broad set of visual concepts. However, they lack expert knowledge specific to culturally significant architecture (as we show in our comparisons). In this work, we incorporate domain-specific knowledge to adapt an image segmentation model conditioned on free text to our setting; we do this by leveraging weak image-level text supervision and pixel-level supervision obtained from multi-view  correspondences. Additionally, we later lift this semantic understanding to volumetric probabilities over a neural representation of the scene.

\smallskip
\noindent \textbf{Language-grounded scene understanding \new{and exploration}.} 
The problem of 3D visual grounding aims at localizing objects in a 3D scene, which is usually represented as a point cloud~\cite{huang2022multi,chen2022ham,chen2022language, lu2023open}. Several works have exploited free-form language for object localization~\cite{chen2020scanrefer, chen2022d3net} or semantic segmentation~\cite{rozenberszki2022language} of a 3D scene provided as an RGB-D scan. Peng \etal~\cite{peng2022openscene} have leveraged input images in addition to a 3D model, represented as a mesh or a point cloud, to co-embed dense 3D point features with image pixels and natural language. 

These works generally assume strong supervision from existing semantically annotated 3D data, consisting of common standalone objects. By contrast, we tackle the challenging real-world scenario of a photo collection in the wild, aiming to localizing semantic regions in large-scale scenes and lacking annotated ground-truth 3D segmentation data for training.
To overcome this lack of strong ground-truth data, our method distills \textit{both} semantic and spatial information from large-scale Internet image collections with textual metadata, and fuses this knowledge together into a neural volumetric field. 

\new{The problem of visualizing and exploring large-scale 3D scenes depicting tourist landmarks captured \emph{in-the-wild} has been explored by several prior works predating the current deep learning dominated era~\cite{snavely2006photo, snavely2008finding, russell20133d}.} Exactly a decade ago, Russell \etal~\shortcite{russell20133d} proposed 3D Wikipedia for annotating isolated 3D reconstructions of famous tourist sites using reference text via image--text co-occurrence statistics. Our work, in contrast, does not assume access to text describing the landmarks of interest and instead leverages weakly-related textual information of similar landmarks. More recently, Wu \etal~\shortcite{wu2021towers} also addressed the problem of connecting 3D-augmented Internet image collections to semantics. %
However, like most prior work, they focused on learning a small set of predefined semantic categories, associated with isolated points in space. By contrast, we operate in the more challenging setting of open-vocabulary semantic understanding, aiming to associate these semantics with volumetric probabilities. %

\smallskip
\noindent \textbf{NeRF-based semantic representations.} 
Recent research efforts have aimed to augment neural radiance fields (NeRF)~\cite{mildenhall2020nerf} with semantic information for segmentation and editing~\cite{turki2023suds}. One approach is to add a classification branch to assign each pixel with a semantic label, complementing the color branch of a vanilla NeRF~\cite{zhi2021in-place, kundu2022panoptic, siddiqui2022panoptic,fu2022panoptic}. %
A general drawback of these categorical methods is the confinement of the segmentation to a pre-determined set of classes.

To enable open-vocabulary segmentation, an alternative approach predicts an \textit{entire feature vector} for each 3D point~\cite{tschernezki22neural, kobayashi2022decomposing, fan2022nerf, kerr2023lerf}; these feature vectors can then be probed with the embedding of a semantic query such as free text or an image patch. While these techniques allow for more flexibility than categorical methods, they perform an ambitious task---regressing high-dimensional feature vectors in 3D space---and are usually demonstrated in controlled capture settings (e.g. with images of constant illumination).

To reduce the complexity of 3D localization for unconstrained large-scale scenes captured in the wild, we adopt a hybrid approach. Specifically, our semantic neural field is optimized over a single text prompt at a time, rather than learning general semantic features which could match arbitrary queries. This enables open-vocabulary segmentation, significantly outperforming alternative methods in our setting.

\footnotetext{\emph{Colonnade} refers to a row of columns separated from each other by an equal distance. A \emph{tympanum} is the semi-circular or triangular decorative wall surface over an entrance, door or window, which is bounded by a lintel and an arch.} 

\section{Method} \label{sec:method}

\begin{figure}
\jsubfig{\fcolorbox{cyan}{cyan}{\includegraphics[height=1.75cm]{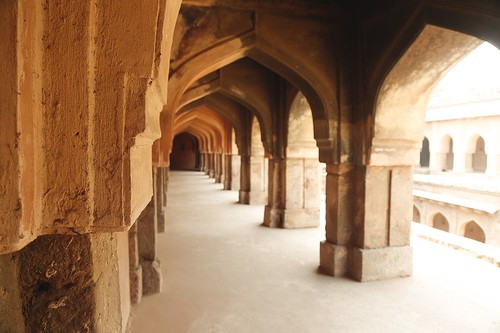}}}{}
\hfill
\jsubfig{\fcolorbox{orange}{orange}{\includegraphics[height=1.75cm]{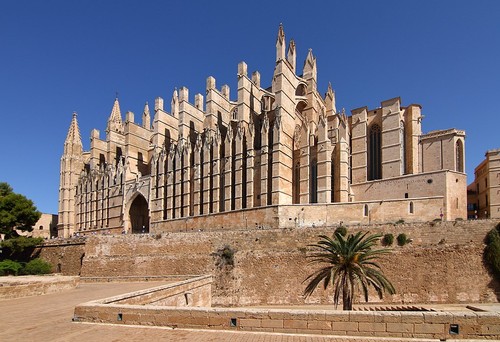}}}{} 
\hfill
\jsubfig{\fcolorbox{teal}{teal}{\includegraphics[height=1.75cm]{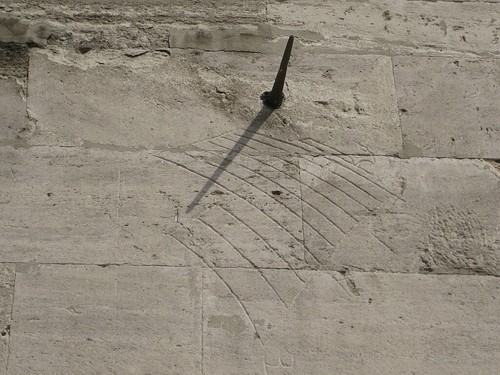}}}{} 
\\ \vspace{-7pt}
  {\begin{flushleft} \leftskip=0.1pt   \footnotesize{
  \archway{
  \textbf{Input}: 
  \textsc{Arched-walkways-at Rajon-ki-Baoli.jpg} ``This is a photo of ASI monument number.'' \texttt{Rajon ki Baoli}. \\
  \textbf{Output}: Archways
  }
  \\
  \facade{
  \textbf{Input:}
  \textsc{Catedral-de-Palma-de-Mallorca,-fachada-sur,-desde-el-Paseo-de-la-Muralla.jpg} ``Catedral de Palma de Mallorca, fachada sur, desde el Paseo de la Muralla.'' \texttt{mallorca catedral cathedral palma spain mallorca majorca;Exterior of Cathedral of Palma de Mallorca;Cathedral of Palma de Mallorca - Full}. 
  \\
  \textbf{Output}: Facade
  }
  \\
  \sundial{
  \textbf{Input:}
  \textsc{Sundial-yeni camii2-istanbul.jpg} ``sundial outside Yeni Camii. On top of the lines the arabic word Asr (afternoon daily prayer) is given. The ten lines (often they are only 9) indicate the times from 20min to 3h before the prayer. Time is read off at the tip of the shadow. The clock was made around 1669 (1074 H).'' \texttt{New Mosque (Istanbul)}. \\
  \textbf{Output}: Sundial}}
  \end{flushleft}} %
  \vspace{-3pt}
  \caption{
  \textbf{LLM-based distillation of semantic concepts}. The full image metadata (\textbf{Input}), including \textsc{Filename}, ``\emph{caption}'' and \texttt{\emph{WikiCategories}} (depicted similarly above) are used for extracting distilled semantic pseudo-labels (\textbf{Output}) with an LLM. Note that the associated images on top (depicted with corresponding colors) are not used as inputs for the computation of their pseudo-labels.
  }\label{fig:distilled}
\end{figure}

An overview of the proposed system is presented in Figure \ref{fig:system}. Our goal is to perform text-driven neural 3D localization for landmark scenes captured by collections of Internet photos. In other words, given this collection of images and a text prompt describing a semantic concept in the scene (for example, \emph{windows} or \emph{spires}), we would like to know where it is located in 3D space.
These images are \emph{in the wild}, meaning that they may be taken in different seasons, time of day, viewpoints, and distances from the landmark, and may include transient occlusions. 

In order to localize unique architectural features landmarks in 3D space, we leverage the power of modern foundation models for visual and textual understanding. Despite progress in general multimodal understanding, modern VLMs struggle to localize fine-grained semantic concepts on architectural landmarks, as we show extensively in our results. The architectural domain uses a specialized vocabulary, with terms such as \textit{pediment} and \textit{tympanum} being rare in general usage; furthermore, terms such as \textit{portal} may have a particular domain-specific meaning in architecture (referring primarily to doors) in contrast to its general usage (meaning any kind of opening). 

To address these challenges, we design a three-stage system: the \emph{offline} stages of LLM-based semantic concept distillation (Section \ref{sec:distill}) and semantic adaptation of VLMs (Section \ref{sec:semantic-adapt}), followed by the \emph{online} stage of 3D localization (Section \ref{sec:nerf}). In the offline stages of our method, we learn relevant semantic concepts using textual metadata as guidance by distilling it via an LLM, and subsequently locate these concepts in space by leveraging inter-view correspondences. The resulting fine-tuned image segmentation model is then used in the online stage to supervise the learning of volumetric probabilities---associating regions in 3D space with the probability of depicting the target text prompt.

\paragraph*{Training Data}The training data for learning the unique semantics of such landmarks is provided by the WikiScenes dataset~\cite{wu2021towers}, consisting of images capturing nearly one hundred \emph{Cathedrals}. We augment these with images capturing 734 \emph{Mosques}, using their data scraping procedure\footnote{Unlike \cite{wu2021towers} that only use images of more common landmarks that can also be reconstructed using \emph{structure-from-motion} techniques, we also include landmarks that are captured by several images only.}. We also remove all landmarks used in our \ourdataset{} benchmark (described in Section \ref{sec:benchmark}) from this training data to prevent data leakage. The rich data captured in both textual and visual modalities in this dataset, along with large-scale coverage of a diverse set of scenes, provides the needed supervision for our system.

\subsection{LLM-Based Semantic Concept Distillation} \label{sec:distill}
In order to associate images with relevant semantic categories for training, we use their accompanying textual metadata as weak supervision. As seen in Figure \ref{fig:distilled}, this metadata is highly informative but also noisy, often containing many irrelevant details as well as having diverse formatting and multilingual contents.
Prior work has shown that such data can be distilled into categorical labels that provide a supervision signal~\cite{wu2021towers}; however, this loses the long tail of uncommon and esoteric categories which we are interested in capturing.
Therefore, we leverage the power of instruction-tuned large language models (LLMs) for distilling concise, open-ended semantic \emph{pseudo-labels} from image metadata using an instruction alone (i.e. zero-shot, with no ground-truth supervision). In particular, we use the encoder-decoder LLM Flan-T5~\cite{chung2022scaling}, which performs well on tasks requiring short answers and is publicly available (allowing for reproducibility of our results). To construct a prompt for this model, we concatenate together the image's filename, caption, and WikiCategories (\emph{i.e.}, a hierarchy of named categories provided in Wikimedia Commons) into a single description string; we prepend this description with the instruction: ``\emph{What architectural feature of \B is described in the following image? Write "unknown" if it is not specified.}'' In this prompt template, the building's name is inserted in \B (e.g. \emph{Cologne Cathedral}). We then generate a pseudo-label using beam search decoding, and lightly process these outputs with standard textual cleanup techniques. Out of ${\sim}101K$ images with metadata in our train split of WikiScenes, this produces ${\sim}58K$ items with non-empty pseudo-labels (those passing filtering heuristics), consisting of 4,031 unique values. Details on text generation settings, textual cleanup heuristics, and further statistics on the distribution of pseudo-labels are provided in the supplementary material. 

Qualitatively, we observe that these pseudo-labels succeed in producing concise English pseudo-labels for inputs regardless of distractor details and multilingual data. This matches the excellent performance of LLMs such as Flan-T5 on similar tasks such as text summarization and translation. Several examples of the metadata and our generated pseudo-labels are provided in Figure \ref{fig:distilled}, and a quantitative analysis of pseudo-label quality is given in our ablation study (Section \ref{sec:ablation}).

\begin{figure}
\rotatebox{90}{\textit{... quire screen\setcounter{footnote}{2}\protect\footnotemark{} }}
    \vspace{3pt}
  \jsubfig{\includegraphics[height=2.05cm]{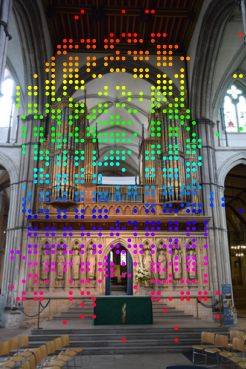} \hfill \includegraphics[height=2.05cm]{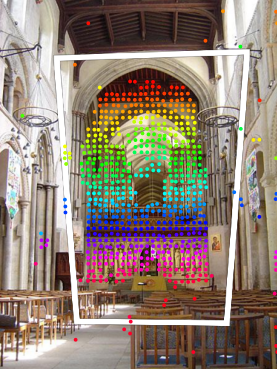}}{ }
 \hfill 
  \jsubfig{\includegraphics[height=2.05cm]{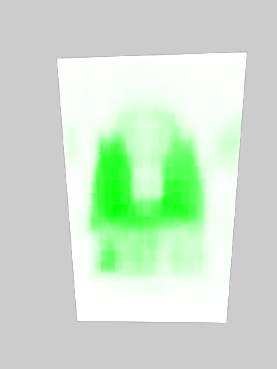}}{  }
  \hfill
   \jsubfig{\includegraphics[height=2.05cm]{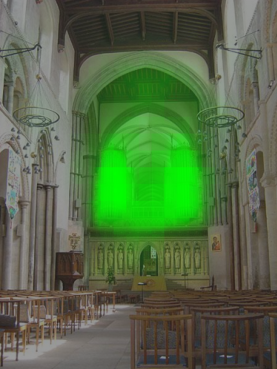} \hfill \includegraphics[height=2.05cm]{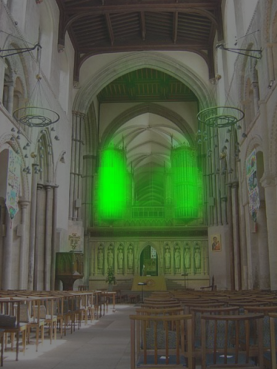}}{  }
   \\ \vspace{3pt}
\rotatebox{90}{\whitetxt{xxpx}\textit{Pulpit}}\hfill %
  \jsubfig{\includegraphics[height=2.05cm]{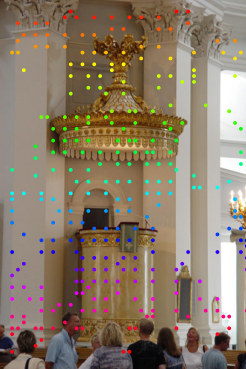} \hfill \includegraphics[height=2.05cm]{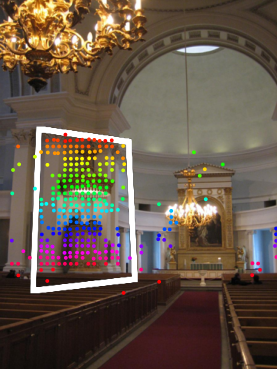}}{ }
 \hfill 
  \jsubfig{\includegraphics[height=2.05cm]{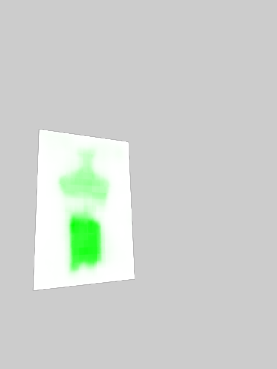}}{  }
  \hfill
   \jsubfig{\includegraphics[height=2.05cm]{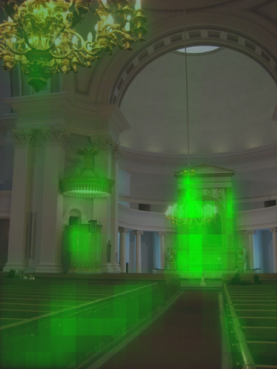} \hfill \includegraphics[height=2.05cm]{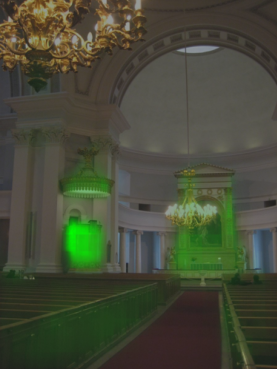}}{  }
   \\ \vspace{3pt}
\rotatebox{90}{\whitetxt{x}\textit{Dome}} \hfill
  \jsubfig{\includegraphics[height=1.03cm]{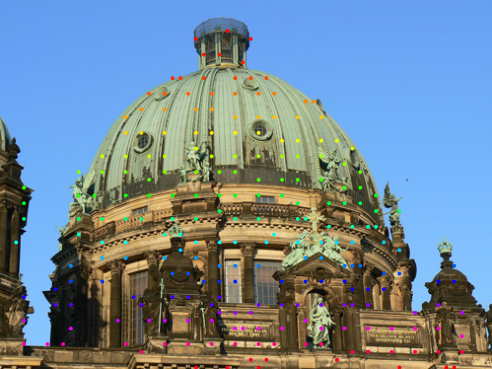} \hfill \includegraphics[height=1.55cm]{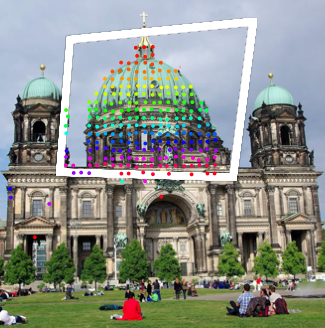}}{ \small{Corresponding images}}
 \hfill 
  \jsubfig{\includegraphics[height=1.55cm]{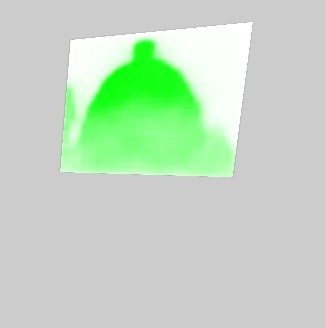}}{ \small{Supervision}}
  \hfill
   \jsubfig{\includegraphics[height=1.55cm]{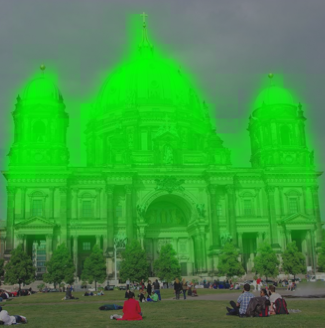} \hfill \includegraphics[height=1.55cm]{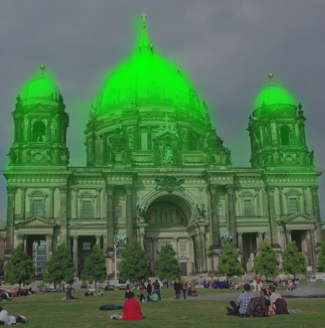}}{  \small{Before \& After fine-tuning}}
    \caption{
  \textbf{Adapting a text-based image segmentation model to architectural landmarks}. We utilize image correspondences (such as the pairs depicted on the left) and pseudo-labels to fine-tune CLIPSeg. We propogate the pseudo-label and pseudo-label of the zoomed-in image to serve as the supervision target, as shown in the central column; we supervise predictions on the zoomed-out image only over the corresponding region (other regions are colored in grayed out for illustration purposes).
  This supervision (together with using random crops further described in the text) refines the model's ability to recognize and localize architectural concepts, as seen by the improved performance shown on the right.
  }\label{fig:adaptation}
\end{figure}

\subsection{Semantic Adaptation of V\&L Models}
\label{sec:semantic-adapt}

After assigning textual pseudo-labels to training images as described in Section \ref{sec:distill}, we use them as supervision for cross-modal understanding, learning image-level and pixel-level semantics. As we show below in Section \ref{sec:experiments}, existing V\&L models lack the requisite domain knowledge out of the box, struggling to understand architectural terms or to localize them in images depicting large portions of buildings. We therefore adapt pretrained models to our setting, using image-pseudolabel pairs to learn \emph{image-level} semantics and weak supervision from pairwise image correspondences to bootstrap \emph{pixel-level} semantic understanding. We outline the training procedures of these models here; see the supplementary material for further details.

To learn image-level semantics of unique architectural concepts in our images, we fine-tune the popular foundation model CLIP~\cite{radford2021learning}, a dual encoder model pretrained with a contrastive text-image matching objective. This model encodes images and texts in a shared semantic space, with cross-modal similarity reflected by cosine distance between embeddings. Although CLIP has impressive zero-shot performance on many classification and retrieval tasks, it may be fine-tuned on text-image pairs to adapt it to particular semantic domains. We fine-tune with the standard contrastive learning objective using our pairs of pseudo-labels and images, and denote the resulting refined model by \ourclip{}. \new{In addition to being used for further stages in our VLM adaptation pipeline, \ourclip{} serves to retrieve relevant terminology for the user who may not be familiar with architectural terms, as we show in our evaluations (Section \ref{sec:qual}).}

To apply our textual pseudo-labels and image-level semantics to concept localization, we build on the recent segmentation model CLIPSeg~\cite{luddecke2022image}, which allows for zero-shot text-conditioned image segmentation. CLIPSeg uses image and text features from a CLIP backbone along with additional fusion layers in an added decoder component; trained on text-supervised segmentation data, this shows impressive open-vocabulary understanding on general text prompts. While pretrained CLIPSeg fails to adequately understand architectural concepts or to localize them (as we show in Section \ref{sec:ablation}), it shows a basic understanding of some concepts along with a tendency to attend to salient objects (as we further illustrate in the supplementary material), which we exploit to bootstrap understanding in our setting.

Our key observation is that large and complex images are composed of subregions with different semantics (e.g. the region around a window or portal of a building), and pretrained CLIPSeg predictions on these zoomed-in regions are closer to the ground truth than predictions on the entire building facade.
To find such pairs of zoomed-out and zoomed-in images, we use two types of geometric connections: multi-view geometric correspondences (i.e. \emph{between images}) and image crops (i.e. \emph{within images}). Using these paired images and our pseudo-label data, we use predictions on zoomed-in views as supervision to refine segmentation on zoomed-out views.

For training across multiple images, we use a feature matching model~\cite{sun2021loftr} to find robust geometric correspondences between image pairs and \ourclip{} to select pairs where the semantic concept (given by a pseudo-label) is more salient in the zoomed-in view relative to the zoomed-out view; for training within the same image, we use \ourclip{} to select relevant crops. We use pretrained CLIPSeg to segment the salient region in the zoomed-in or cropped image, and then fine-tune CLIPSeg to produce this result in the relevant image when zoomed out; we denote the resulting trained model by \ourclipseg{}. During training we freeze CLIPSeg's encoders, training its decoder module alone with loss functions optimizing for the following:

\footnotetext{Full pseudo-label text: \textit{Neo-gothic quire screen}. This refers to a screen that partitions the choir (or quire) and the aisles in a cathedral or a church. }

\noindent \textbf{Geometric correspondence supervision losses.} As described above, we use predictions on zoomed-in images to supervise segmentation of zoomed-out views. We thus define loss terms $\mathcal{L}_{corresp}$ and $\mathcal{L}_{crop}$, the cross-entropy loss of these predictions calculated on the region with supervision targets, for correspondence-based and crop-based data respectively. \new{In other words, $\mathcal{L}_{corresp}$ encourages predictions on zoomed-out images to match predictions on corresponding zoomed-in views as seen in Figure \ref{fig:adaptation}; $\mathcal{L}_{crop}$ is similar but uses predictions on a crop of the zoomed-out view rather than finding a distinct image with a corresponding zoomed-in view.}

\noindent \textbf{Multi-resolution consistency.}  To encourage consistent predictions across resolutions and to encourage our model to attend to relevant details in all areas of the image, we use a multi-resolution consistency loss $\mathcal{L}_{consistency}$ calculated as follows. Selecting a random crop of the image from the correspondence-based dataset, we calculate cross-entropy loss between our model's prediction cropped to this region, and CLIPSeg (pretrained, without fine-tuning) applied within this cropped region. To attend to more relevant crops, we pick the random crop by sampling two crops from the given image and using the one with higher \ourclip{} similarity to the textual pseudo-label.

\noindent \textbf{Regularization.} We add the regularization loss $\mathcal{L}_{reg}$, calculated as the average binary entropy of our model's outputs. This encourages confident outputs (probabilities close to $0$ or $1$).

These losses are summed together with equal weighting; further training settings, hyperparameters, and data augmentation are detailed in the supplementary material.

\begin{figure}
\rotatebox{90}{\whitetxt{xxx}Input}
  \jsubfig{\includegraphics[height=1.70cm]{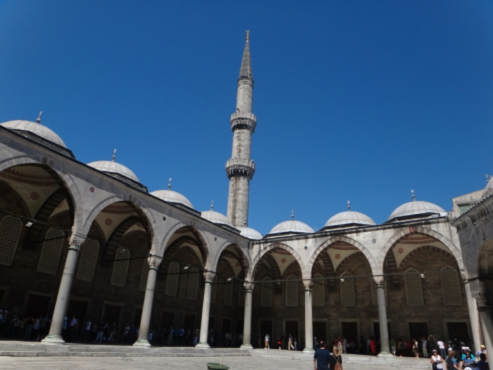}}{  }
  \hfill
   \jsubfig{\includegraphics[height=1.70cm]{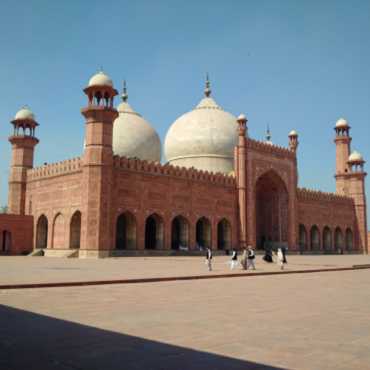}}{  }
   \hfill
    \jsubfig{\includegraphics[height=1.70cm]{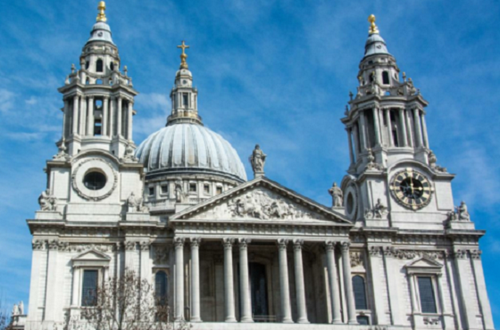}}{  }
       \hfill
    \jsubfig{\includegraphics[height=1.70cm]{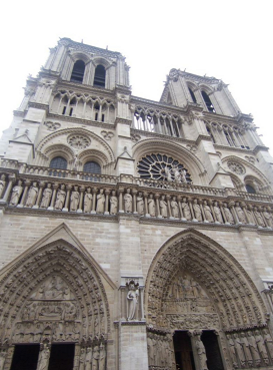}}{  }
 \\
\rotatebox{90}{\whitetxt{xxp}Before}
  \jsubfig{\includegraphics[height=1.70cm]{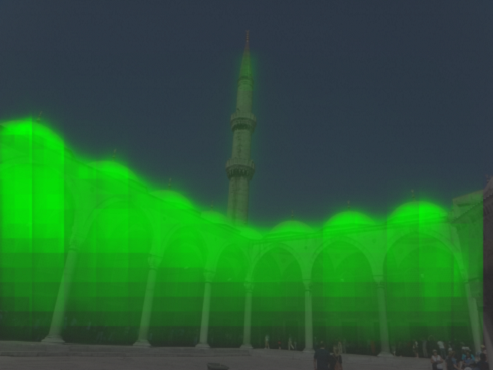}}{  }
  \hfill
   \jsubfig{\includegraphics[height=1.70cm]{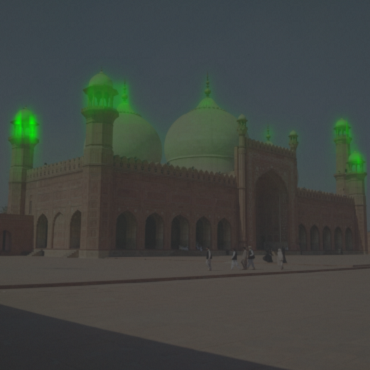}}{  }
   \hfill
    \jsubfig{\includegraphics[height=1.70cm]{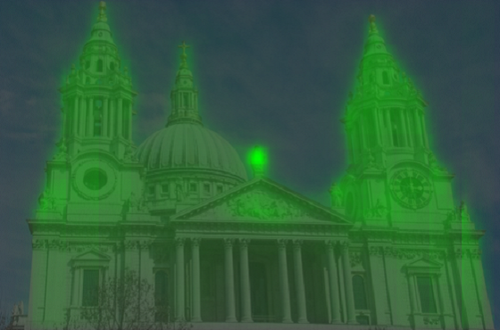}}{  }
       \hfill
    \jsubfig{\includegraphics[height=1.70cm]{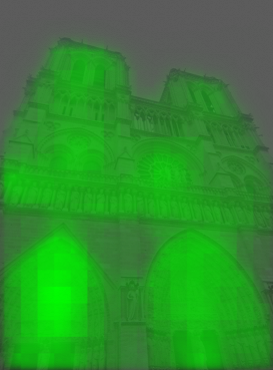}}{  }
 \\
 \rotatebox{90}{\whitetxt{xxp}After}
  \jsubfig{\includegraphics[height=1.70cm]{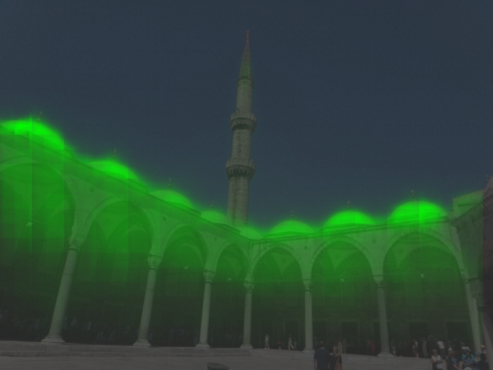}}{ \textit{Domes} }
  \hfill
   \jsubfig{\includegraphics[height=1.70cm]{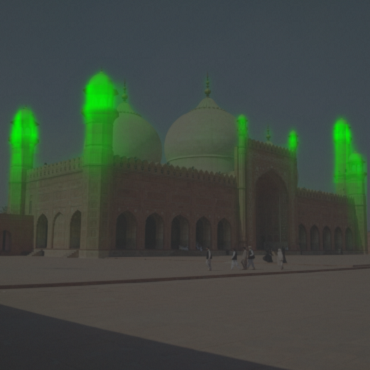}}{ \textit{Minarets} }
   \hfill
    \jsubfig{\includegraphics[height=1.70cm]{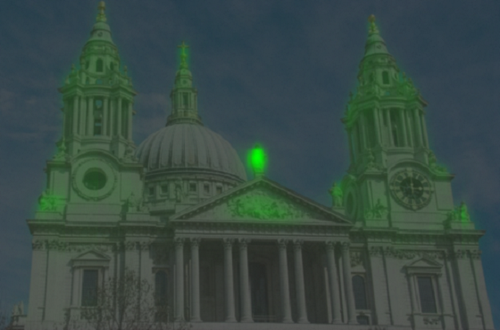}}{\textit{Statues}  }
   \hfill
    \jsubfig{\includegraphics[height=1.70cm]{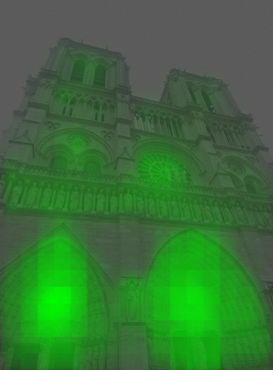}}{\textit{Tympanum}  }
    \caption{
  \textbf{Text-based segmentation before and after fine-tuning}. Above we show 2D segmentation results over images belonging to landmarks from \ourdataset{} (unseen during training). As illustrated above, our weakly-supervised fine-tuning scheme improves the segmentation of domain-specific semantic concepts.
  }\label{fig:adaptation_test}
\end{figure}

We illustrate this fine-tuning process over corresponding image pairs in Figure \ref{fig:adaptation}. As illustrated in the figure, the leftmost images (\emph{i.e.}, zoom-ins) determine the supervision signal. Note that while we only supervise learning in the corresponding region in each training sample, the refined model (denoted as \ourclipseg{}) correctly extrapolates this knowledge to the rest of the zoomed-out image. Figure \ref{fig:adaptation_test} illustrates the effect of this fine-tuning on segmentation of new landmarks (unseen during training); we see that our fine-tuning gives \ourclipseg{} knowledge of various semantic categories that the original pretrained CLIPSeg struggles to localize; we proceed to use this model to produce 2D segmentations that may be lifted to a 3D representation.

\subsection{Text-Driven Neural 3D Localization} \label{sec:nerf}
In this section, we describe our approach for performing 3D localization over a neural representation of the scene, using the semantic understanding obtained in the previous offline training stages. The input to our 3D localization framework is an Internet image collection of a new (unseen) landmark and a target text prompt.

First, we optimize a Ha-NeRF~\cite{chen2022hallucinated} representation to learn volumetric densities and colors from the unstructured image collection. We then extend this neural representation by adding a semantic output channel. Inspired by previous work connecting neural radiance fields with semantics~\cite{zhi2021in-place}, we augment Ha-NeRF with a segmentation MLP head, added on top of a shared backbone (see the supplementary material for additional details). To learn the volumetric probabilities of given target text prompt, we freeze the shared backbone and optimize only the segmentation MLP head. 

To provide supervision for semantic predictions, we use the 2D segmentation map predictions of \ourclipseg{} (described in Section \ref{sec:semantic-adapt}) on each input view. While these semantically adapted 2D segmentation maps are calculated for each view separately, \ourmethod{} learns a 3D model which aggregates these predictions while enforcing 3D consistency. We use a binary cross-entropy loss to optimize the semantic volumetric probabilities, comparing them to the 2D segmentation maps over sampled rays \cite{zhi2021in-place}.
This yields a a representation of the semantic concept's location in space. Novel rendered views along with estimated probabilities are shown in Figures \ref{fig:teaser} and \ref{fig:system} and in the accompanying videos.

\section{The \ourdataset{} Benchmark} \label{sec:benchmark}

\begin{figure}
\centering
    \jsubfig{\includegraphics[height=2.6cm]{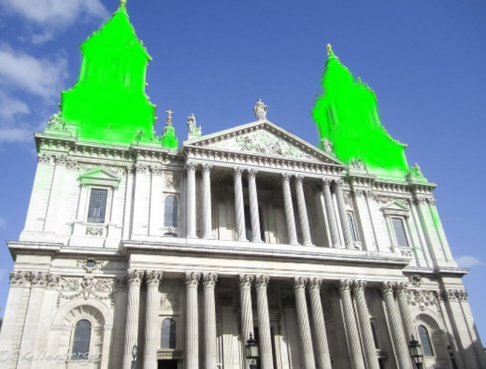}}{\textit{Towers}}
    \hfill
    \jsubfig{\includegraphics[height=2.6cm]{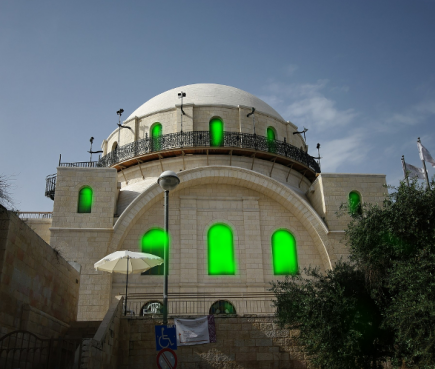}}{\textit{Windows}}
    \hfill
    \jsubfig{\includegraphics[height=2.6cm]{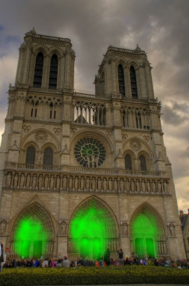}}{\textit{Portals}}

    \vspace{4pt}
    \jsubfig{\includegraphics[height=1.52cm]{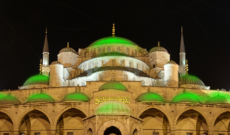}}{\textit{Domes}}
    \hfill
    \jsubfig{\includegraphics[height=1.52cm]{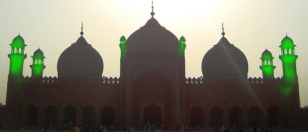}}{\textit{Minarets}}
    \hfill
    \jsubfig{\includegraphics[height=1.52cm]{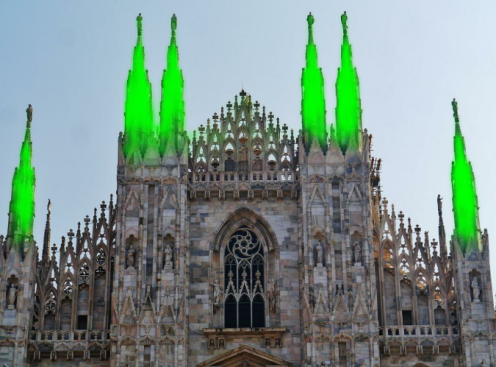}}{\textit{Spires}} 
    
\caption{\textbf{Neural 3D Localization Results.} We show results from each landmark in our \ourdataset{} benchmark (clockwise from top: St. Paul's Cathedral, Hurva Synagogue, Notre-Dame Cathedral, Blue Mosque, Badshahi Mosque, Milan Cathedral), visualizing segmentation maps rendered from 3D \ourmethod{} representations on input scene images. As seen above, \ourmethod{} succeeds in localizing various semantic concepts across diverse landmarks.
}
\label{fig:localization}
\end{figure}

To evaluate our method, we need Internet photo collections covering scenes, paired with ground truth segmentation maps. As we are not aware of any such existing datasets, we introduce the \emph{\ourdataset{}} benchmark, assembled from multiple datasets (WikiScenes \cite{wu2021towers}, IMC-PT 2020~\cite{yi2020image} MegaDepth \cite{li2018megadepth}) along with additional data collected using the data scraping procedure of Wu \etal. We enrich these scene images with ground-truth segmentation annotations. Our dataset includes 6,305 images associated with 3D structure-from-motion reconstructions and ground-truth segmentations for multiple semantic categories. 

We select six landmarks, exemplified in Figure \ref{fig:localization}: \emph{Notre-Dame Cathedral} (Paris), \emph{Milan Cathedral} (Milan), \emph{St. Paul's Cathedral} (London), \emph{Badshahi Mosque} (Lahore), \emph{Blue Mosque} (Istanbul) and \emph{Hurva Synagogue} (Jerusalem). These landmarks span different geographical regions, religions and characteristics, and can readily be associated with accurate 3D reconstructions due to the large number of publicly-available Internet images. We associate these landmarks with the following semantic categories: \textit{portal}, \textit{window}, \textit{spire}, \textit{tower}, \textit{dome}, and \textit{minaret}. Each landmark is associated with a subset of these categories, according to its architectural structure (\emph{e.g.}, \textit{minaret} is only associated with the two mosques in our benchmark). 

We produce ground-truth segmentation maps to evaluate our method using manual labelling combined with correspondence-guided propagation. For each semantic concept, we first manually segment several images from different landmarks. We then propagate these segmentation maps to overlapping images, and manually filter these propagated masks (removing, for instance, occluded images). Additional details about our benchmark are provided in the supplementary material.

\section{Results and Evaluation} \label{sec:experiments}

In this section, we evaluate the performance of \ourmethod{} on the \ourdataset{} benchmark, and compare our method to recent works on text-guided semantic segmentation and neural localization techniques. We also validate each component of our system with ablation studies  -- namely, our LLM-based concept distillation, VLM semantic adaptation, and 3D localization. Finally, we discuss limitations of our approach. \new{In the supplementary material, we provide experimental details as well as additional experiments, such as an evaluation of the effect of CLIPSeg fine-tuning on general and architectural term understanding evaluated on external datasets.}

\subsection{Baselines} \label{sec:baselines}
We compare our method to text-driven image segmentation methods, as well as 3D NeRF segmentation techniques. As \ourdataset{} consists of paired images and view-consistent segmentation maps, it can be used to evaluate both 2D and 3D segmentation methods; in the former case, by directly segmenting images and evaluating on their ground-truth (GT) annotations; in the latter case, by rendering 2D segmentation masks from views corresponding to each GT annotation.

For text-based 2D segmentation baseline methods, we consider CLIPSeg~\cite{luddecke2022image} and LSeg~\cite{li2022language}. We also compare to the ToB model proposed by Wu \etal~\shortcite{wu2021towers} that learns image segmentation over the WikiScenes dataset using cross-view correspondences as weak supervision. As their model is categorical, operating over only ten categories, we report the performance of ToB only over the semantic concepts included in their model.

For 3D NeRF-based segmentation methods, we consider DFF~\cite{kobayashi2022decomposing} and LERF~\cite{kerr2023lerf}. Both of these recent methods utilize text for NeRF-based 3D semantic segmentation. DFF~\cite{kobayashi2022decomposing} performs semantic scene decomposition using text prompts, distilling text-aligned image features into a volumetric 3D representation and segmenting 3D regions by probing these with the feature representation of a given text query. Similarly, LERF optimizes a 3D language field from multi-scale CLIP embeddings with volume rendering.

The publicly available implementations of DFF and LERF cannot operate on our \emph{in-the-wild} problem setting, as it does not have images with constant illumination or a single camera model. To provide a fair comparison, we replace the NeRF backbones used by DFF and LERF (vanilla NeRF and Nerfacto respectively) with Ha-NeRF, as used in our model, keeping the remaining architecture of these models unchanged. \new{In the supplementary material, we also report results over the unmodified DFF and LERF implementations using constant illumination images rendered from Google Earth.}

\new{In addition to these existing 3D methods, we compare to the baseline approach of lifting 2D CLIPSeg (pretrained, not fine-tuned) predictions to a 3D representation with Ha-NeRF augmented with a localization head (as detailed in Section \ref{sec:nerf}). This baseline, denoted as HaLo-NeRF-, provides a reference point for evaluating the relative contribution of our optimization-based approach rather than learning a feature field which may be probed for various textual inputs (as used by competing methods), and of our 2D segmentation fine-tuning.}

\begin{table}[t]
  \centering
  \setlength{\tabcolsep}{3.2pt}
  \begin{tabularx}{0.99\columnwidth}{@{ } llccccccc @{}}
    \toprule
    Method & \footnotesize{mAP} & \footnotesize{portal} & \footnotesize{window} & \footnotesize{spire} & \footnotesize{tower} & \footnotesize{dome} & \footnotesize{minaret} \\
    \midrule
    \textbf{2D Seg.} \\
    LSeg & 0.09 & 0.05 & 0.13 & 0.06 & 0.19 & 0.05 & 0.06\\
    ToB & 0.23 & 0.15 & 0.04 & $\times$ & 0.49 & $\times$ & $\times$\\
    CLIPSeg & 0.56 & 0.29 & 0.44 & 0.46 & 0.87 & 0.69 & 0.63 \\
    \hl{\ourclipseg{}} & \textbf{0.66} & \textbf{0.49} & \textbf{0.51} & \textbf{0.50} & \textbf{0.87} & \textbf{0.77} & \textbf{0.81} \\
    \midrule
    \textbf{3D Loc. } \\
    DFF$^*$  & 0.11 & 0.06 & 0.04 & 0.09 & 0.17 & 0.11 & 0.17\\
    LERF$^*$  & 0.14 & 0.16 & 0.15 & 0.18 & 0.13 & 0.10 & 0.09\\
    \new{HaLo-NeRF-} & 0.62 & 0.28 & 0.61 & 0.55 & \textbf{0.90} & 0.72 & 0.69 \\ %
    \hl{HaLo-NeRF} & \textbf{0.68} & \textbf{0.45} & \textbf{0.64} & \textbf{0.61} & 0.87 & \textbf{0.74} & \textbf{0.80} \\
    \bottomrule
  \end{tabularx}
{\begin{flushleft}
  \footnotesize $^*$Using a Ha-NeRF backbone $\qquad$ \new{-Using CLIPSeg without fine-tuning}  %
\end{flushleft}}
\vspace{-2pt}
  \caption{\textbf{Quantitative Evaluation}. We report mean average precision (mAP; averaged per category) and per category average precision over the \ourdataset{} benchmark, comparing our results (\hl{highlighted} in the table) to 2D segmentation and 3D localization techniques. Note that ToB uses a categorical model, and hence we only report performance over concepts it was trained on. Best results are highlighted in \textbf{bold}.}
\label{tab:results}
\end{table}

\subsection{Quantitative Evaluation} \label{sec:evaluation}

As stated in Section \ref{sec:baselines}, our benchmark allows us to evaluate segmentation quality for both both 2D and 3D segmentation methods, in the latter case by projecting 3D predictions onto 2D views with ground-truth segmentation maps. We perform our evaluation using pixel-wise metrics relative to ground-truth segmentations. Since we are interested in the quality of the model's soft probability predictions, we use average precision (AP) as our selected metric as it is threshold-independent.

In Table \ref{tab:results} we report the AP per semantic category (averaged over landmarks), as well as the overall mean AP (mAP) across categories. We report results for 2D image segmentation models on top, and 3D segmentation methods underneath. In addition to reporting 3D localization results for our full proposed system, we also report the results of our intermediate 2D segmentation component (\ourclipseg{}).

As seen in the table, \ourclipseg{} \new{(our fine-tuned segmentation model, as defined in Section 3.2)} outperforms other 2D methods, showing better knowledge of architectural concepts and their localization. In addition to free-text guided methods (LSeg and CLIPSeg), we also outperform the ToB model (which was trained on WikiScenes), consistent with the low recall scores reported by Wu \etal~\shortcite{wu2021towers}. LSeg also struggles in our free-text setting where semantic categories strongly deviate from its training data; CLIPSeg shows better zero-shot understanding of our concepts out of the box, but still has a significance performance gap relative to \ourclipseg{}.

In the 3D localization setting, we also see that our method strongly outperforms prior methods over all landmarks and semantic categories. \ourmethod{} adds 3D-consistency over \ourclipseg{} image segmentations, further boosting performance by fusing predictions from multi-view inputs into a 3D representation which enforces consistency across observations. \new{We also find an overall performance boost relative to the baseline approach using \ourmethod{} without CLIPSeg fine-tuning. This gap is particularly evident in unique architectural terms such as \emph{portal} and \emph{minaret}.}

Regarding the gap between our performance and the competing 3D methods (DFF, LERF), we consider multiple contributing factors. In addition to our enhanced understanding of domain-specific semantic categories and their positioning, the designs of these models differ from \ourmethod{} in ways which may impact performance. DFF is built upon LSeg as its 2D backbone; hence, its performance gap on our benchmark follows logically from the poor performance of LSeg in this setting (as seen in the reported 2D results for LSeg), consistent with the observation of Kobayashi \etal~\shortcite{kobayashi2022decomposing} that DFF inherits bias towards in-distribution semantic categories from LSeg (e.g. for traffic scenes). LERF, like DFF, regresses a full semantic 3D feature field which may then be probed for arbitrary text prompts. By contrast, \ourmethod{} optimizes for the more modest task of localizing a particular concept in space, likely more feasible in this challenging setting. \new{The significant improvement provided by performing per-concept optimization is also supported by the relatively stronger performance of the baseline model shown in Table \ref{tab:results}, which performs this optimization using pretrained (not fine-tuned) CLIPSeg segmentation maps as inputs.} 

\subsection{Qualitative Results} \label{sec:qual}

\begin{figure}
\centering %
\rotatebox{90}{\whitetxt{xxxxxx}\textit{Towers}}
\hspace{1pt}
\jsubfig{\includegraphics[trim={0 0 0 0}, clip, width=2.6cm]{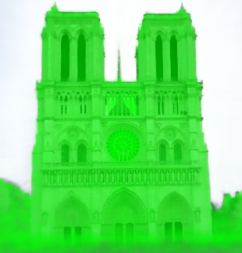}}{}
\hfill
\jsubfig{\includegraphics[trim={0 0 0 0}, clip, width=2.6cm]{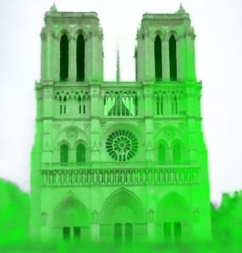}}{}
\hfill
\jsubfig{\includegraphics[trim={0 0 0 0}, clip, width=2.6cm]{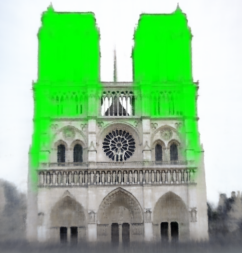}}{}
\rotatebox{90}{\whitetxt{xx}\textit{Windows}}
\hspace{1pt}
\jsubfig{\includegraphics[trim={0 0 0 0}, clip, width=2.6cm]{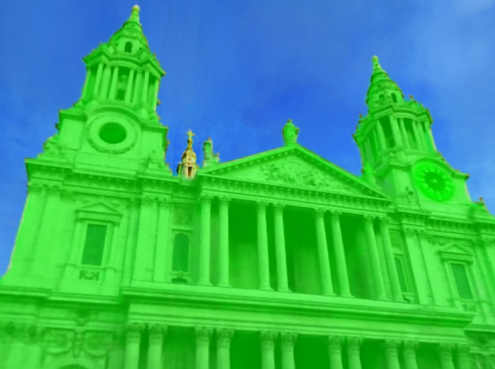}}{}
\hfill
\jsubfig{\includegraphics[trim={0 0 0 0}, clip, width=2.6cm]{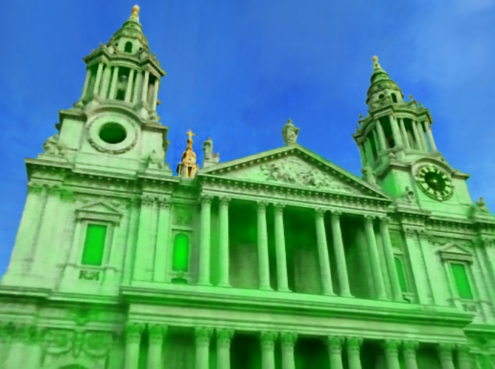}}{}
\hfill
\jsubfig{\includegraphics[trim={0 0 0 0}, clip, width=2.6cm]{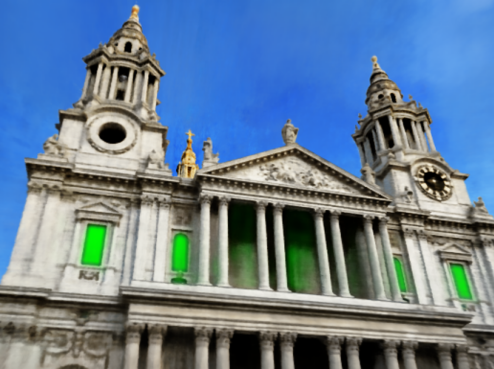}}{}
\rotatebox{90}{\whitetxt{x}\textit{Portals}}
\hspace{1pt}
\jsubfig{\includegraphics[trim={0 0 0 0}, clip, width=2.6cm]{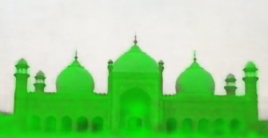}}{DFF$^*$}
\hfill
\jsubfig{\includegraphics[trim={0 0 0 0}, clip, width=2.6cm]{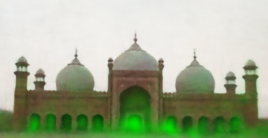}}{LERF$^*$}    
\hfill 
\jsubfig{\includegraphics[trim={0 0 0 0}, clip, width=2.6cm]{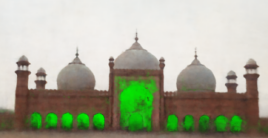}}
{Ours}
\\ %
\vspace{4pt}
{\begin{flushleft}
  \footnotesize $^*$Using a Ha-NeRF backbone
\end{flushleft}}
\vspace{-4pt}
\caption{\textbf{Localizing semantic regions in architectural landmarks compared to prior work.} We show probability maps for DFF and LERF models on \emph{Milan Cathedral}, along with our results. As seen above, DFF and LERF struggle to distinguishing between different semantic regions on the landmark, while our method accurately localizes the semantic concepts.}
\label{fig:nerf_comp}
\end{figure}

\begin{figure}
\centering
    \rotatebox{90}{\;\;\;\textit{Baseline}}
    \jsubfig{\includegraphics[height=1.83cm]{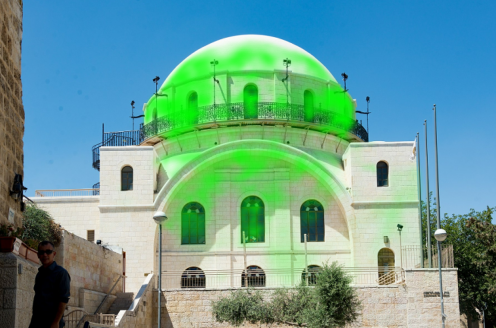}}{\textit{}}
    \hfill
    \jsubfig{\includegraphics[height=1.83cm]{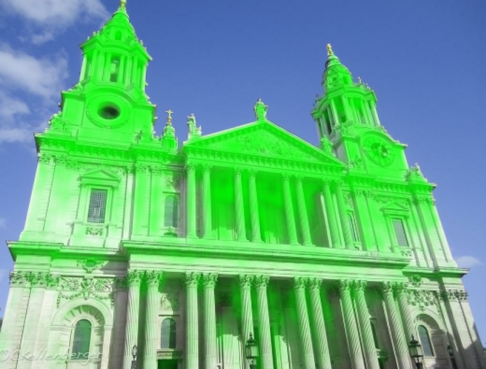}}{\textit{}}
    \hfill
    \jsubfig{\includegraphics[height=1.83cm]{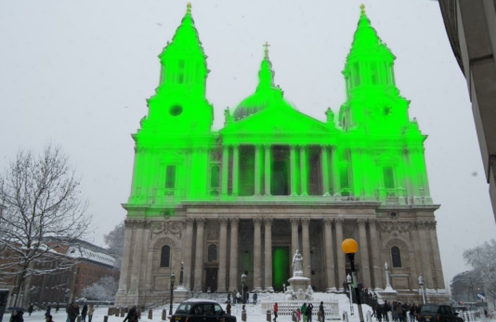}}{\textit{}} 
   \\
    \vspace{-8pt}
    \rotatebox{90}{\;\;\;\;\;\;\textit{Ours}}
    \jsubfig{\includegraphics[height=1.83cm]{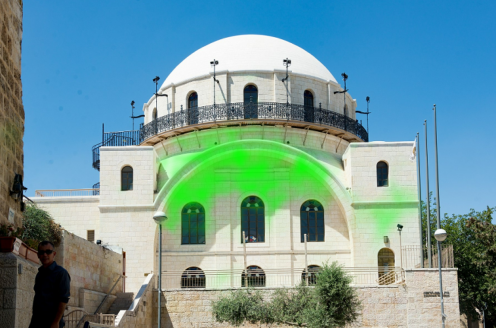}}{\textit{Arch}}
    \hfill
    \jsubfig{\includegraphics[height=1.83cm]{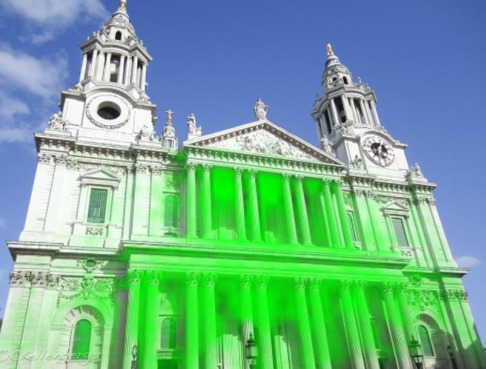}}{\textit{Colonnade}}
    \hfill
    \jsubfig{\includegraphics[height=1.83cm]{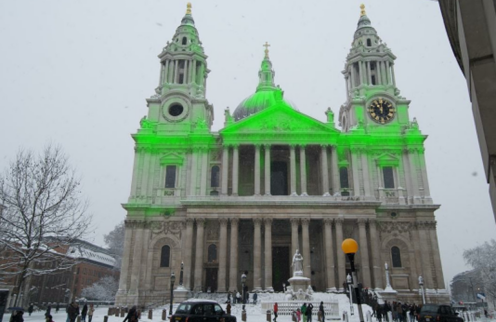}}{\textit{Pediment}} 
\caption{\textbf{3D localization results on additional concepts}, comparing \ourmethod{} to the baseline HaLo-NeRF- model (using CLIPSeg without fine-tuning as input to \ourmethod) over semantic regions appearing on the \emph{Hurva Synagogue} (left) and \emph{St. Paul's Cathedral} (right). Our model can localize these concepts, while the baseline model fails to reliably distinguish between relevant and irrelevant regions. See below for the definitions of the concepts shown\setcounter{footnote}{3}\protect\footnotemark{}.}
\label{fig:ablation}
\end{figure}

\begin{figure}
\centering
    \jsubfig{\includegraphics[height=2.18cm]{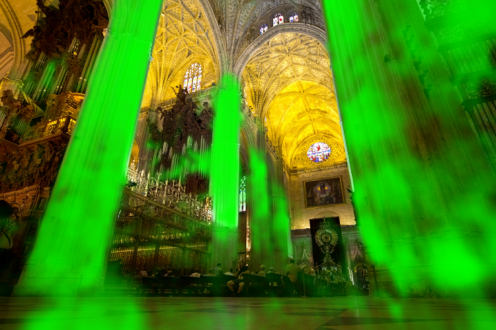}}
    {\textit{Pillars}}
    \hfill
    \jsubfig{\includegraphics[height=2.18cm]{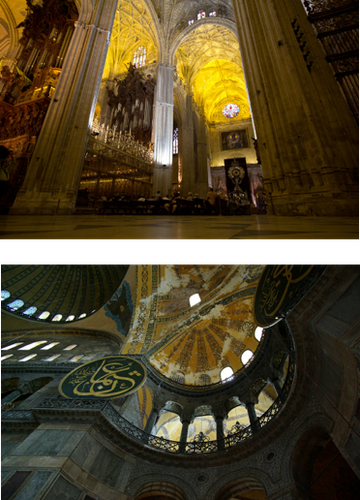}}
    {\textit{}}
    \hfill
    \jsubfig{\includegraphics[height=2.18cm]{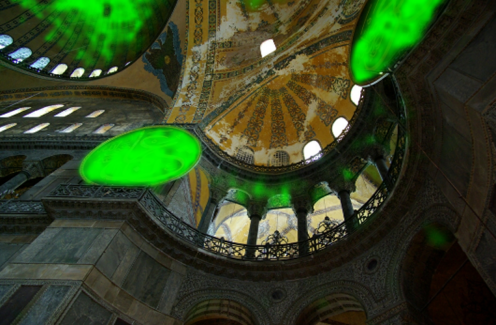}}
    {\textit{Roundel}}
\caption{\textbf{Results over indoor scenes.} \ourmethod{} is capable of localizing unique semantic regions within building interiors (shown above over the \emph{Seville Cathedral} and \emph{Blue Mosque} landmarks). The definition of \emph{roundel} is given below\setcounter{footnote}{4}\protect\footnotemark{}. %
}
\label{fig:indoor}
\end{figure}

\begin{figure}
    \begin{minipage}{0.25\linewidth}
        \vspace{0pt}
        \fcolorbox{cyan}{cyan}{\includegraphics[trim={0cm 8cm 0cm 8cm},clip,width=\linewidth]{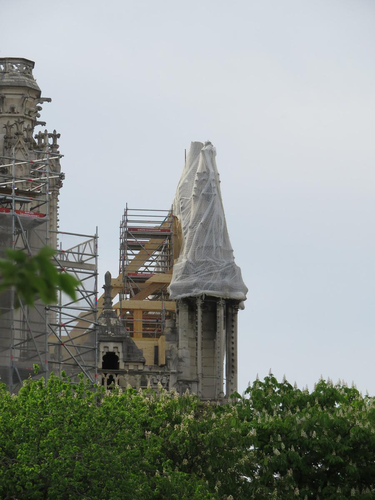}}
        \fcolorbox{orange}{orange}{\includegraphics[trim={0cm 0cm 0cm 0cm},clip,width=\linewidth]{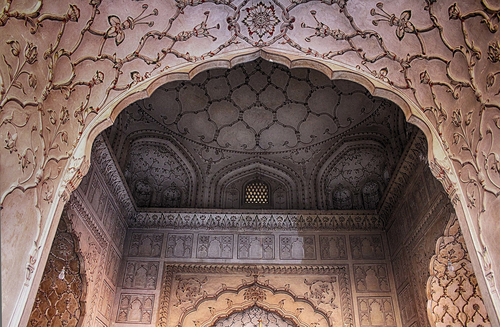}}
    \end{minipage}
    \hfill
    \begin{minipage}{0.7\linewidth}
        \begin{tabularx}{\linewidth}[t]{C|C}
            \multicolumn{2}{c}{\textbf{\ourclip{} Img$\to$Text Results}} \\
            \rowcolor{cyan!25} \emph{construction} & \cellcolor{orange!25} \emph{muqarnas} \\
            \rowcolor{cyan!25} \emph{scaffolding} & \cellcolor{orange!25} \emph{ornate} \\
            \rowcolor{cyan!25} \emph{bell tower}
            & \cellcolor{orange!25} \emph{stucco decoration} \\
            \rowcolor{cyan!25} \emph{crossing tower}
            & \cellcolor{orange!25} \emph{dome chamber} \\
            \rowcolor{cyan!25} \emph{church tower}
            & \cellcolor{orange!25} \emph{tile work} \\
            \rowcolor{cyan!25} \emph{clock tower}
            & \cellcolor{orange!25} \emph{mihrab} \\
            \rowcolor{cyan!25} \emph{sail tower}
            & \cellcolor{orange!25} \emph{ceiling tile work} \\
            \rowcolor{cyan!25} \emph{flying buttresses}
            & \cellcolor{orange!25} \emph{winter prayer hall}
        \end{tabularx}
    \end{minipage}
    \caption{\new{\textbf{Examples of terminology retrieval.} By applying \ourclip{} to a given view, the user may retrieve relevant architectural terminology which can then be localized with \ourmethod{}. Above, we display the top eight retrieval results for two test images, using the \ourclip{} retrieval methodology described in Section \ref{sec:qual}. As is seen above, \ourclip{} returns relevant items such as \emph{scaffolding}, \emph{church tower}, \emph{muqarnas}, \emph{ceiling tile work} which may aid the user in selecting relevant architectural terms.}}
    \label{fig:retrieval}
\end{figure}

\begin{figure*}
\centering
    \rotatebox{90}{\;\;\;\textit{Quadriga}}
    \jsubfig{\includegraphics[height=1.83cm]{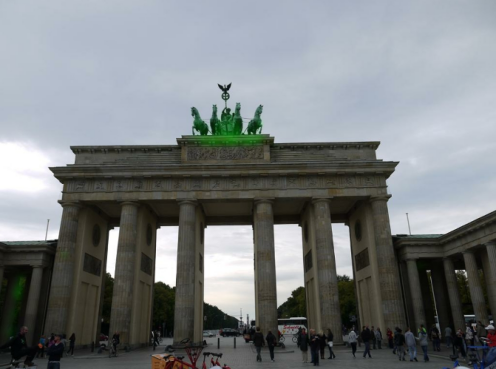}}
    {}
    \jsubfig{\includegraphics[height=1.83cm]{figures/images_zoom_ins/quadriga_0190.jpg}}
    {\textit{}}
    \jsubfig{\includegraphics[height=1.83cm]{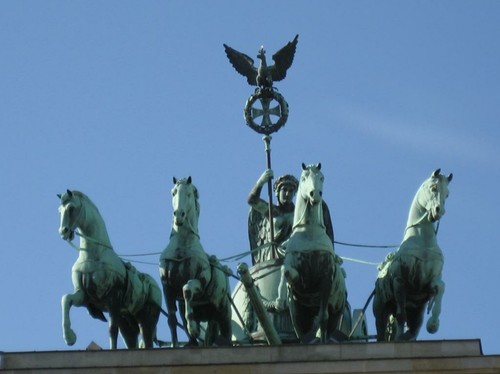}}
    {\textit{}}
    \jsubfig{\includegraphics[height=1.83cm]{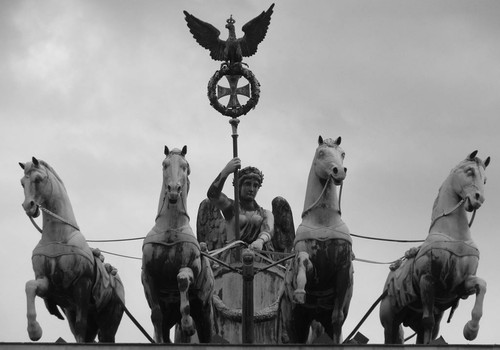}}
    {\textit{}}
    \hfill
    \rotatebox{90}{\;\;\;\;\;\;\textit{Clock}}
    \jsubfig{\includegraphics[height=1.83cm]{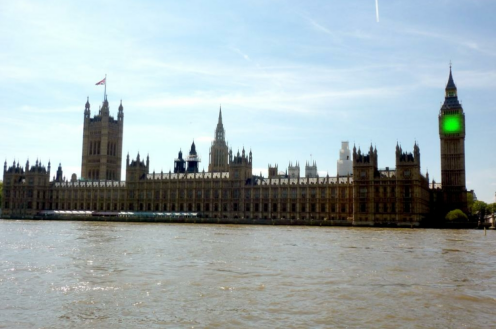}}
    {}
    \jsubfig{\includegraphics[trim={0 0 400 0}, clip, height=1.83cm]{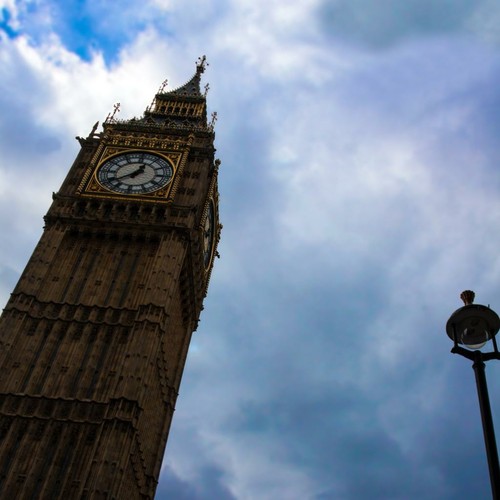}}
    {\textit{}}
    \jsubfig{\includegraphics[height=1.83cm]{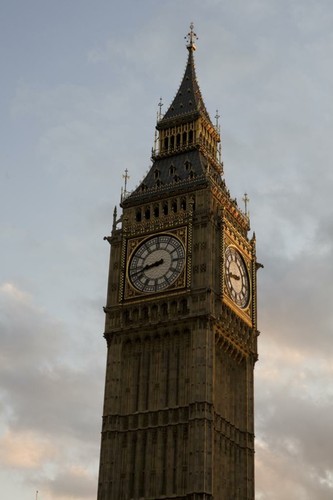}}
    {\textit{}}
    \jsubfig{\includegraphics[height=1.83cm]{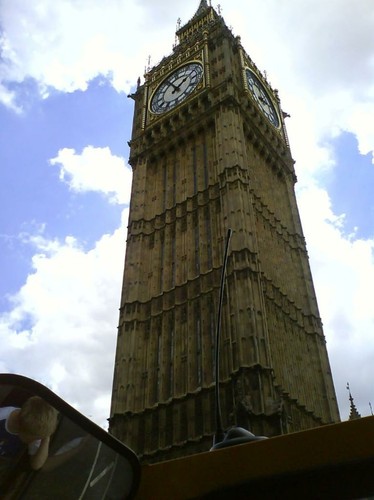}}
    {\textit{}}

    \rotatebox{90}{\;\;\;\;\;\textit{\new{Glass}}}
    \jsubfig{\includegraphics[height=1.64cm]{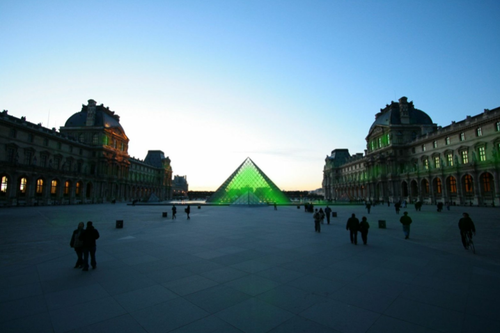}}
    {}
    \jsubfig{\includegraphics[height=1.64cm]{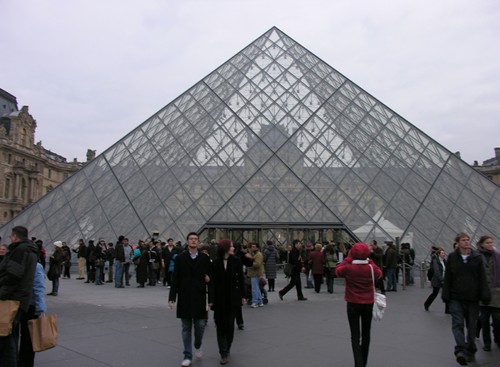}}
    {\textit{}}
    \jsubfig{\includegraphics[height=1.64cm]{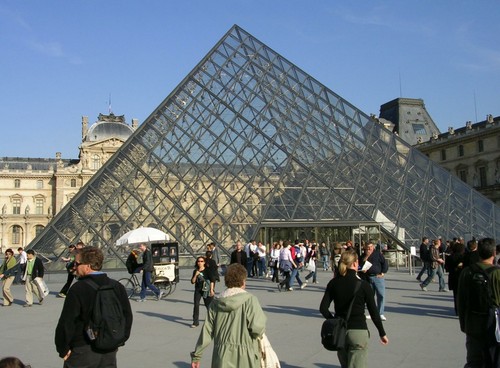}}
    {\textit{}}
    \jsubfig{\includegraphics[height=1.64cm]{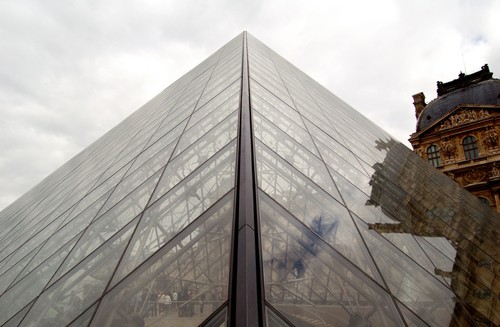}}
    {\textit{}}
    \hfill
    \rotatebox{90}{\;\;\;\textit{\new{The Egg}}}
    \jsubfig{\includegraphics[height=1.64cm]{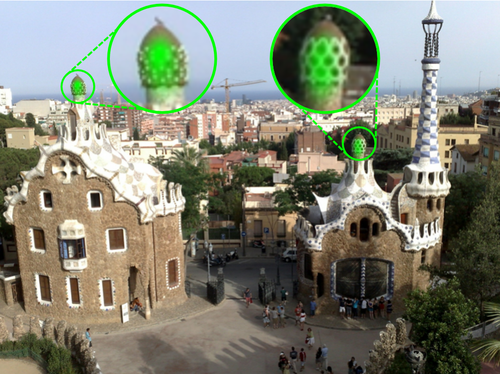}}
    {\textit{}}
    \jsubfig{\includegraphics[height=1.64cm]{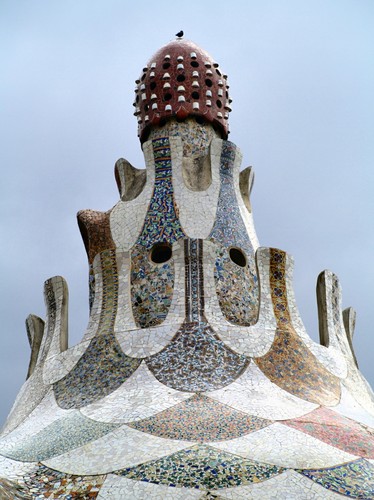}}    {\textit{}}
    \jsubfig{\includegraphics[height=1.64cm]{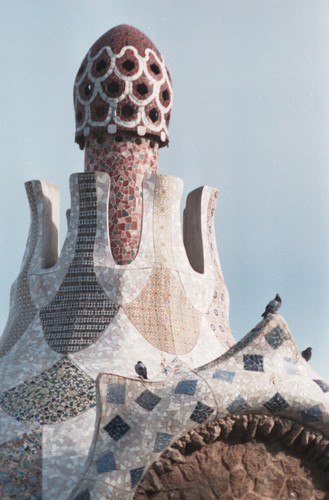}}
    {\textit{}}
    \jsubfig{\includegraphics[height=1.64cm]{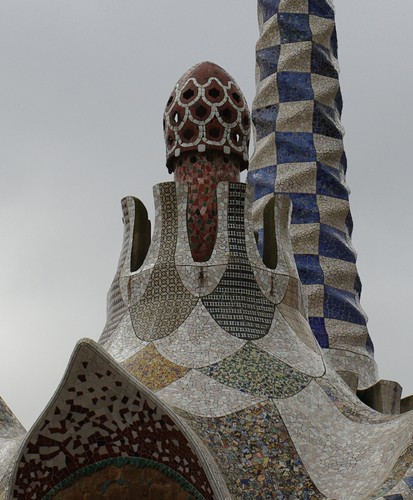}}
    {\textit{}}
    \jsubfig{\includegraphics[height=1.64cm]{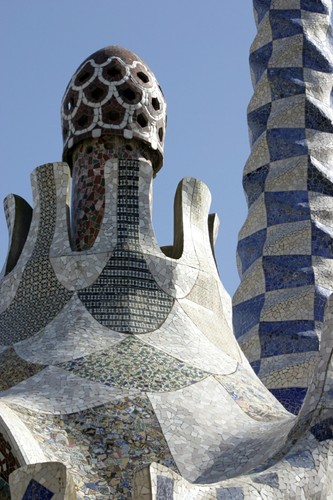}}
    {\textit{}}

    \rotatebox{90}{\;\;\;\;\;\;\;\textit{\new{Torch}}}
    \jsubfig{\includegraphics[height=1.85cm]{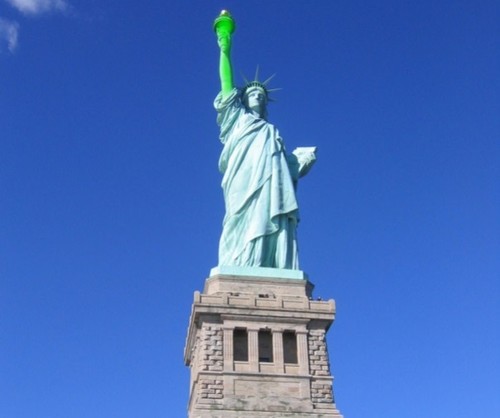}}{}
    \jsubfig{\includegraphics[height=1.85cm]{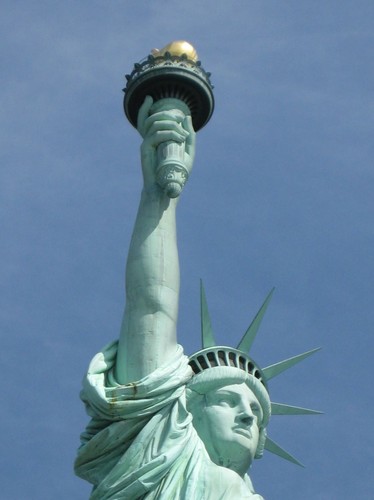}}
    {\textit{}}
    \jsubfig{\includegraphics[height=1.85cm]{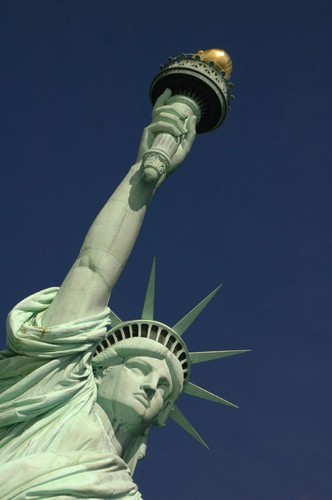}}
    {\textit{}}
    \jsubfig{\includegraphics[height=1.85cm]{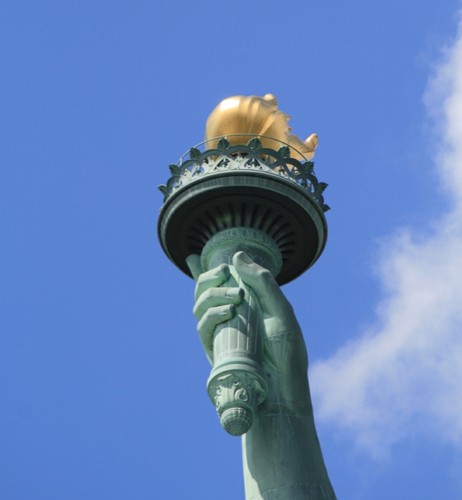}}
    {\textit{}}
    \jsubfig{\includegraphics[height=1.85cm]{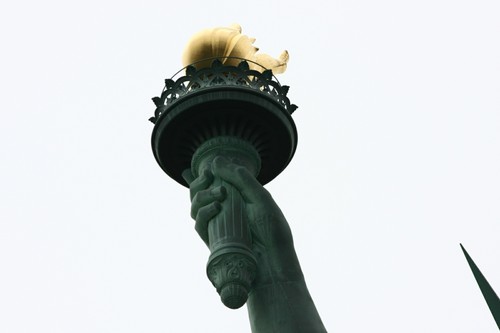}}
    {\textit{}}
    \hfill
    \rotatebox{90}{\;\;\;\textit{\new{The Eiffel}}}
    \jsubfig{\includegraphics[height=1.85cm]{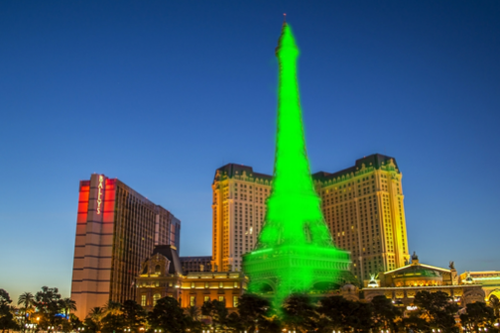}}
    {\textit{}}
     \jsubfig{\includegraphics[height=1.85cm]{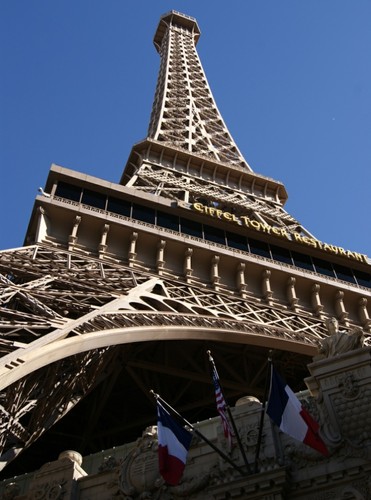}}
    {\textit{}}
    \jsubfig{\includegraphics[height=1.85cm]{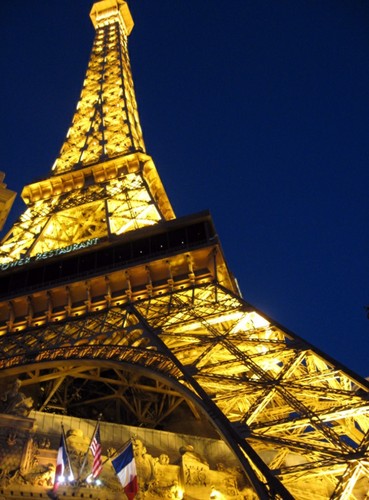}}
    {\textit{}}
    \jsubfig{\includegraphics[height=1.85cm]{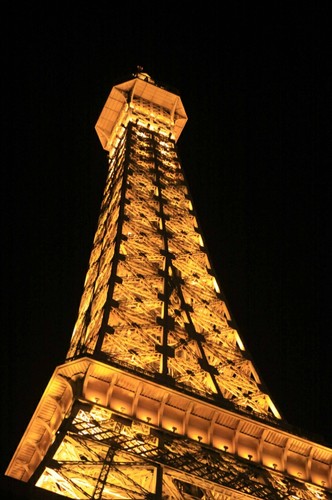}}
    {\textit{}}

    \rotatebox{90}{\;\;\;\;\;\;\textit{Arch}}
    \jsubfig{\includegraphics[height=1.56cm]{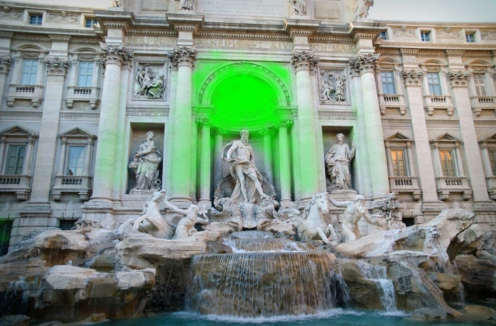}}
    {}
    \jsubfig{\includegraphics[height=1.56cm]{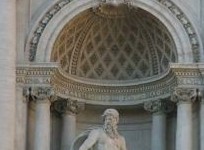}}
    {\textit{}}
    \jsubfig{\includegraphics[height=1.56cm]{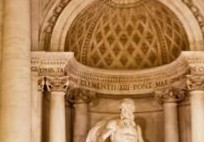}}
    {\textit{}}
    \jsubfig{\includegraphics[height=1.56cm]{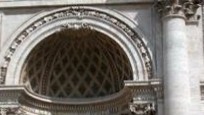}}
    {\textit{}}
    \hfill
    \rotatebox{90}{\;\;\;\;\;\;\textit{Text}}
    \jsubfig{\includegraphics[height=1.56cm]{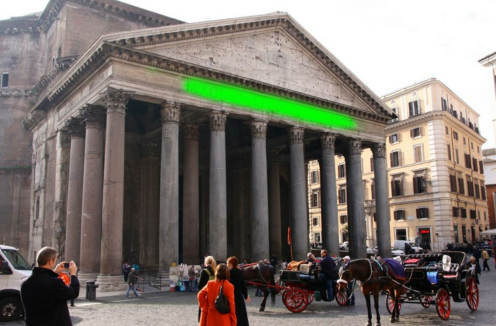}}
    {}
    \jsubfig{\includegraphics[height=1.56cm]{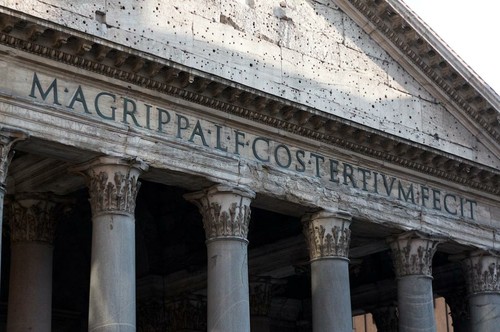}}
    {\textit{}}
    \jsubfig{\includegraphics[height=1.56cm]{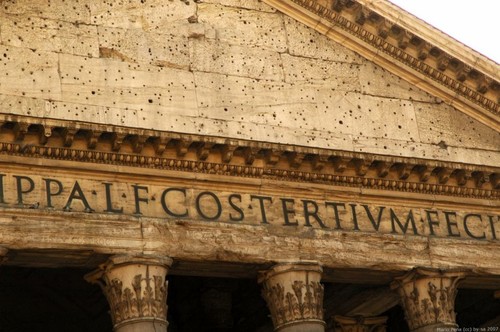}}
    {\textit{}}

    \rotatebox{90}{\;\;\;\:\textit{Statues}}
    \jsubfig{\includegraphics[height=1.675cm]{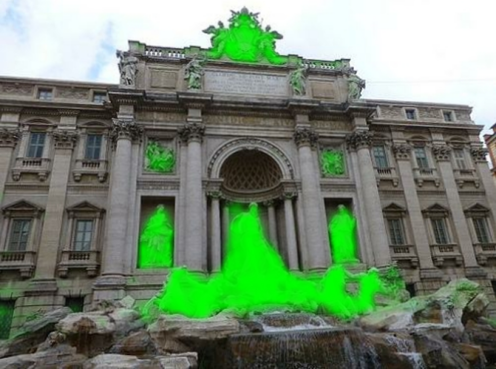}}
    {}
    \jsubfig{\includegraphics[height=1.675cm]{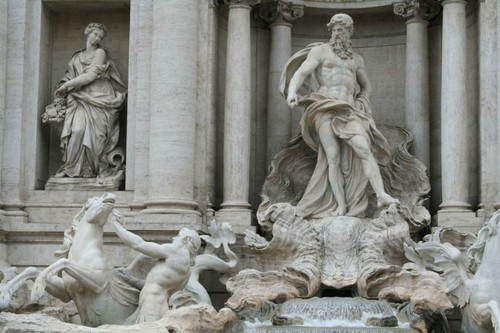}}
    {\textit{}}
    \jsubfig{\includegraphics[height=1.675cm]{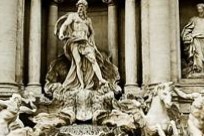}}
    {\textit{}}
    \jsubfig{\includegraphics[height=1.675cm]{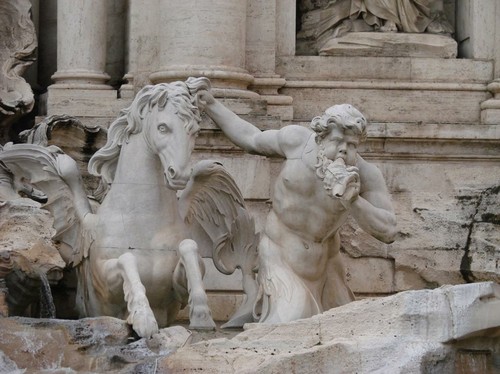}}
    {\textit{}}
    \hfill
    \rotatebox{90}{\;\;\;\;\;\textit{Fence}}
    \jsubfig{\includegraphics[height=1.675cm]{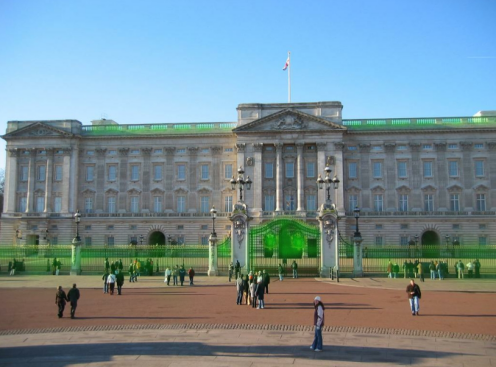}}
    {}
    \jsubfig{\includegraphics[height=1.675cm]{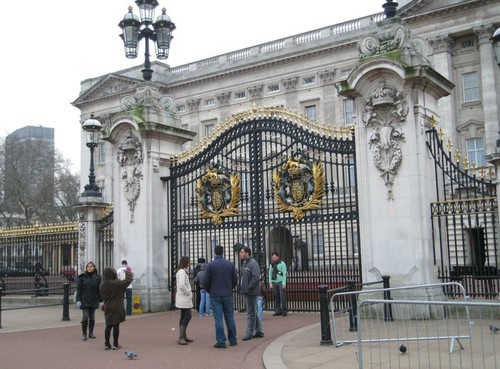}}
    {\textit{}}
    \jsubfig{\includegraphics[height=1.675cm]{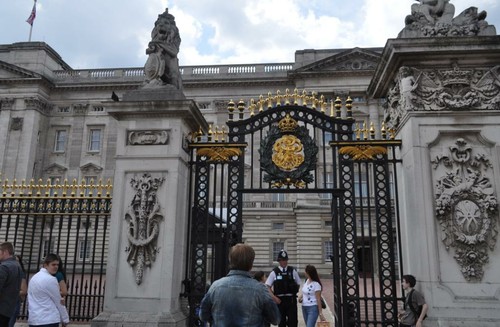}}
    {\textit{}}

\caption{\textbf{Localization for general architectural scenes.} \ourmethod{} can localize various semantic concepts in a variety of scenes in the wild, not limited to the religious domain of \ourdataset{}. Our localization, marked in green in the first image for each concept, enables focusing automatically on the text-specified region of interest, as shown by the following zoomed-in images in each row.
}
\label{fig:outdoor}
\end{figure*}

Sample results of our method are provided in Figures \ref{fig:localization}--\ref{fig:outdoor}. As seen in Figure \ref{fig:localization}, \ourmethod{} segments regions across various landmarks and succeeds in differentiating between fine-grained architectural concepts. Figure \ref{fig:nerf_comp} compares these results to alternate 3D localization methods. As seen there, alternative methods fail to reliably distinguish between the different semantic concepts, tending to segment the entire building facade rather than identifying the areas of interest. With LERF, this tendency is often accompanied by higher probabilities in coarsely accurate regions, as seen by the roughly highlighted windows in the middle row. \new{Figure \ref{fig:ablation} shows a qualitative comparison of \ourmethod{} with and without CLIPSeg fine-tuning over additional semantic concepts beyond those from our benchmark. As is seen there, our fine-tuning procedure is needed to learn reliable localization of such concepts which may be lifted to 3D.}

We include demonstrations of the generality of our method. Besides noting that our test set includes the \textit{synagogue} category which was not seen in training (see the results for the Hurva Synagogue shown in Figure \ref{fig:localization}), we test our model in the more general case of (non-religious) architectural landmarks. Figure \ref{fig:outdoor} shows results on various famous landmarks captured in the IMC-PT 2020 dataset~\cite{yi2020image} (namely, Brandenburg Gate, Palace of Westminster, \new{The Louvre Museum, Park Güell, The Statue of Liberty, Las Vegas}, The Trevi Fountain, The Pantheon, and The Buckingham Palace). As seen there, \ourmethod{} localizes unique scene elements such as the \textit{quadriga} in the Brandenburg Gate, \new{the Statue of Liberty's torch, and the Eiffel tower}, \new{The Statue of Liberty, and Las Vegas, respectively}. \new{In addition, \ourmethod{} localizes} common semantic concepts, such as \textit{clock}, \new{\textit{glass},} and \textit{text} in the Palace of Westminster, \new{The Louvre Museum,} and The Pantheon, respectively. Furthermore, while we focus mostly on outdoor scenes, Figure \ref{fig:indoor} shows that our method can also localize semantic concepts over reconstructions capturing indoor scenes.

\new{Understanding that users may not be familiar with fine-grained or esoteric architectural terminology, we anticipate the use of \ourclip{} (our 
fine-tuned CLIP model, as defined in Section 3.2) for retrieving relevant terminology. In particular, \ourclip{} may be applied to any selected view to retrieve relevant terms to which the user may then apply \ourmethod{}. 
We demonstrate this qualitatively in Figure \ref{fig:retrieval}, which shows the top terms retrieved by \ourclip{} on test images. In the supplementary material, we also report a quantitative evaluation over all architectural terms found at least 10 times in the training data. This evaluation further demonstrates that \ourclip{} can retrieve relevant terms over these Internet images (significantly outperforming pretrained CLIP at this task).}

Figure \ref{fig:outdoor} further illustrates the utility of our method for intuitive exploration of scenes. By retrieving scene images having maximal overlap with localization predictions, the user may focus automatically on the text-specified region of interest, allowing for exploration of the relevant semantic regions of the scene in question. This is complementary to exploration over the optimized neural representation, as illustrated in Figures \ref{fig:teaser}-\ref{fig:system}, and in the accompanying videos.

\begin{table}[t]
  \centering
  \setlength{\tabcolsep}{3.4pt}
  \begin{tabularx}{0.99\columnwidth}{@{ } lccccccccc @{}}
    \toprule
    Method &
    \footnotesize{mAP} & \footnotesize{portal} & \footnotesize{window} & \footnotesize{spire} & \footnotesize{tower} & \footnotesize{dome} & \footnotesize{minaret} \\ 
    \midrule
    \ourclipseg{} & \textbf{0.66} & \textbf{0.49} & \textbf{0.51} & \textbf{0.50} & 0.87 & \textbf{0.77} & \textbf{0.81} \\
    $\;\; -\mathcal{L}_{crop}$ & 0.65 & \textbf{0.49} & 0.50 & 0.47 & \textbf{0.88} & \textbf{0.77} & 0.78 \\
    $\;\; -\mathcal{L}_{reg}$ & 0.63  & 0.45 & 0.46 & 0.45 & 0.87 & \textbf{0.77} & 0.78 \\
    $\;\; -\text{corr. data}^*$ & 0.45  & 0.09 & 0.27 & 0.25 & 0.76 & 0.65 & 0.71 \\
    2D Baseline & 0.56 & 0.29 & 0.44 & 0.46 & 0.87 & 0.69 & 0.63 \\
    \bottomrule
  \end{tabularx}
    \vspace{3pt}
{\begin{flushleft}
  \footnotesize $^*$ Refers to removing correspondence supervision losses, namely $\mathcal{L}_{corresp}$ and $\mathcal{L}_{consistency}$.
\end{flushleft}}
\vspace{3pt}
  \caption{\textbf{Ablation Studies}, evaluating the effect of design choices on the fine-tuning process of \ourclipseg{}. ``Baseline'' denotes using the CLIPSeg segmentation model without fine-tuning. We report AP and mAP metrics over the \ourdataset{} benchmark as in Table \ref{tab:results}. Best results are highlighted in \textbf{bold}.}
\label{tab:ablation}
\end{table}

\setcounter{footnote}{3}\footnotetext{\emph{Colonnade} refers to a row of columns separated from each other by an equal distance. \emph{Pediment} refers to a triangular part at the top of the front of a building that supports the roof and is often decorated.}

\setcounter{footnote}{4}\footnotetext{A \emph{roundel} is an circular shield or figure; here it refers to round panels bearing calligraphic emblems.} 

\subsection{Ablation Studies} \label{sec:ablation}

We proceed to evaluate the contribution of multiple components of our system---LLM-based concept distillation and VLM semantic adaptation---to provide motivation for the design of our full system.

\smallskip \noindent \textbf{LLM-based Concept Distillation}. In order to evaluate the quality of our LLM-generated pseudo-labels and their necessity, we manually review a random subset of 100 items (with non-empty pseudo-labels), evaluating their factual correctness and comparing them to two metadata-based baselines -- whether the correct architectural feature is present in the image's caption, and whether it could be inferred from the last WikiCategory listed in the metadata for the corresponding image (see Section \ref{sec:distill} for an explanation of this metadata). These baselines serve as upper bounds for architectural feature inference using the most informative metadata fields by themselves (and assuming the ability to extract useful labels from them). We find 89\% of pseudo-labels to be factually correct, while only 43\% of captions contain information implying the correct architectural feature, and 81\% of the last WikiCategories to describe said features. We conclude that our pseudo-labels are more informative than the baseline of using the last WikiCategory, and significantly more so than inferring the architectural feature from the image caption. Furthermore, using either of the latter alone would still require summarizing the text to extract a usable label, along with translating a large number of results into English.

\new{To further study the effect of our LLM component on pseudo-labels, we provide ablations on LLM sizes and prompts in the supplementary material, finding that smaller models underperform ours while the best-performing prompts show similar results. There we also} provide statistics on the distribution of our pseudo-labels, showing that they cover a diverse set of categories with a long tail of esoteric items.

\smallskip \noindent \textbf{VLM Semantic Adaptation Evaluation.} To strengthen the motivation behind our design choices of \ourclipseg{}, we provide an ablation study of the segmentation fine-tuning in Table \ref{tab:ablation}. We see that each element of our training design provides a boost in overall performance, together significantly outperforming the 2D baseline segmentation model. In particular, we see the key role of our correspondence-based data augmentation, without which the fine-tuning procedure significantly degrades due to lack of grounding in the precise geometry of our scenes \new{(both relative to full fine-tuning, and relative to the original segmentation model)}.
These results complement Figure \ref{fig:adaptation_test}, which show a qualitative comparison of the CLIPSeg baseline and \ourclipseg{}. \new{We also note that we have provided a downstream evaluation of the effect of fine-tuning CLIPSeg on 3D localization in Table \ref{tab:results}, showing that it provides a significant performance boost and is particularly crucial for less common concepts.}

\subsection{Limitations} \label{sec:limitations}

\new{As our method uses an optimization-based pipeline applied for each textual query, it is limited by the runtime required to fit each term's segmentation field. In particular, a typical run takes roughly two hours on our hardware setup, described in the supplementary material. We foresee future work building upon our findings to accelerate these results, possibly using architectural modifications such as encoder-based distillation of model predictions.}

\begin{figure}
\centering
    \jsubfig{\includegraphics[height=2.75cm]
    {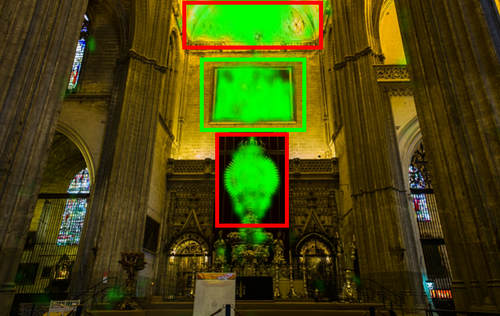}}{\new{\textit{Immaculate Conception}\setcounter{footnote}{5}\protect\footnotemark{}}}
    \hfill
    \jsubfig{\includegraphics[height=2.75cm]{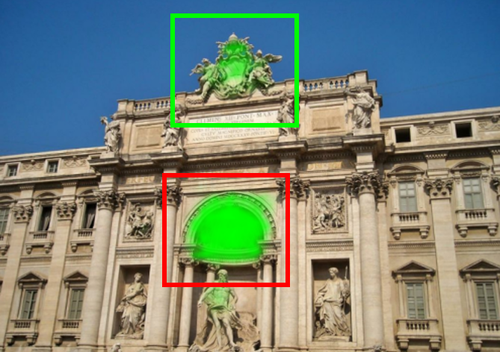}}{\textit{Papal Coat of Arms}} \\
    \vspace{3pt}
    \jsubfig{\includegraphics[height=1.64cm]{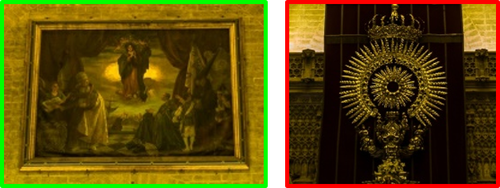}}{}
    \hfill
    \jsubfig{\includegraphics[height=1.64cm]{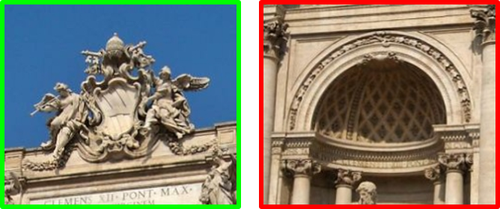}}{} \\
\caption{\textbf{Limitation examples.} Correct results are marked in green boxes and incorrect ones in red. Our method may fail to properly identify terms that never appear in our training data, such as \new{the \emph{Immaculate Conception}\setcounter{footnote}{5}\protect\footnotemark{} as on the left and} the \emph{Papal Coat of Arms} as on the right.}
\label{fig:failure}
\end{figure}

\new{Furthermore, if the user inputs a query which does not appear in the given scene, our model may segment semantically- or geometrically-related regions – behavior inherited from the base segmentation model. For example, the spires of Milan Cathedral are segmented when the system is prompted with the term \emph{minarets}, which are not present in the view but bear visual similarity to spires. Nevertheless, \ourclip{} may provide the user with a vocabulary of relevant terms (as discussed in Section \ref{sec:qual}), mitigating this issue (e.g. \emph{minarets} does not appear among the top terms for images depicting Milan Cathedral). We further discuss this tendency to segment salient, weakly-related regions in the supplementary material.}

Additionally, since we rely on semantic concepts that appear across landmarks in our training set, concepts require sufficient coverage in this training data in order to be learned. While our method is not limited to common concepts and shows understanding of concepts in the long tail of the distribution of pseudo-labels (as analyzed in the supplementary material), those that are extremely rare or never occur in our training data may not be properly identified. This is seen in Figure \ref{fig:failure}, where the localization of the scene-specific concepts \new{\textit{Immaculate Conception} and} \textit{Papal Coat of Arms} (terms which never occur in our training data; for example, the similar term \emph{coat of arms} appears only seven times) incorrectly include other regions.

\setcounter{footnote}{6}\footnotetext{\new{The \emph{Immaculate Conception} is the event depicted in the painting, a work by Alfonso Grosso Sánchez situated in the Seville Cathedral.}}

\section{Conclusions} \label{sec:conclusions}

We have presented a technique for connecting unique architectural elements across different modalities of text, images, and 3D volumetric representations of a scene. To understand and localize domain-specific semantics, we leverage inter-view coverage of a scene in multiple modalities, distilling concepts with an LLM and using view correspondences to bootstrap spatial understanding of these concepts. We use this knowledge as guidance for a neural 3D representation which is view-consistent by construction, and demonstrate its performance on a new benchmark for concept localization in large-scale scenes of tourist landmarks.

Our work represents a step towards the goal of modeling historic and culturally significant sites as explorable 3D models from photos and metadata captured in the wild. We envision a future where these compelling sites are available to all in virtual form, making them accessible and offering educational opportunities that would not otherwise be possible. Several potential research avenues include making our approach interactive, localizing multiple prompts simultaneously and extending our technique to additional mediums with esoteric concepts, such as motifs or elements in artwork. 

\noindent
\textbf{Acknowledgments.} This work was supported by research grants from ISF (application number 2510/23) and BSF (application number 2022363).

\bibliographystyle{eg-alpha-doi} 
\bibliography{references.bib}

\newcommand{\etalchar}[1]{$^{#1}$}
\begin{thebibliography}{\uppercase{WAESS21}}

\bibitem[CCN20]{chen2020scanrefer}
\textsc{Chen D.~Z., Chang A.~X., Nie{\ss}ner M.}:
\newblock {ScanRefer: 3D Object Localization in RGB-D Scans using Natural Language}.
\newblock In \emph{Proceedings of the European Conference on Computer Vision (ECCV)} (2020), pp.~202--221.

\bibitem[CGT{\etalchar{*}}22]{chen2022language}
\textsc{Chen S., Guhur P.-L., Tapaswi M., Schmid C., Laptev I.}:
\newblock Language conditioned spatial relation reasoning for 3d object grounding.
\newblock \emph{arXiv preprint arXiv:2211.09646} (2022).

\bibitem[CHL{\etalchar{*}}22]{chung2022scaling}
\textsc{Chung H.~W., Hou L., Longpre S., Zoph B., Tay Y., Fedus W., Li E., Wang X., Dehghani M., Brahma S., et~al.}:
\newblock {Scaling Instruction-Finetuned Language Models}.
\newblock \emph{arXiv preprint arXiv:2210.11416} (2022).

\bibitem[CLW{\etalchar{*}}22]{chen2022ham}
\textsc{Chen J., Luo W., Wei X., Ma L., Zhang W.}:
\newblock Ham: Hierarchical attention model with high performance for 3d visual grounding.
\newblock \emph{arXiv preprint arXiv:2210.12513} (2022).

\bibitem[CWNC22]{chen2022d3net}
\textsc{Chen D.~Z., Wu Q., Nie{\ss}ner M., Chang A.~X.}:
\newblock {D3Net: A Speaker-Listener Architecture for Semi-supervised Dense Captioning and Visual Grounding in RGB-D Scans}.
\newblock In \emph{Proceedings of the European Conference on Computer Vision (ECCV)} (2022).

\bibitem[CZL{\etalchar{*}}22]{chen2022hallucinated}
\textsc{Chen X., Zhang Q., Li X., Chen Y., Feng Y., Wang X., Wang J.}:
\newblock {Hallucinated Neural Radiance Fields in the Wild}.
\newblock In \emph{Proceedings of the IEEE/CVF Conference on Computer Vision and Pattern Recognition (CVPR)} (2022), pp.~12943--12952.

\bibitem[DLH22]{decatur20223d}
\textsc{Decatur D., Lang I., Hanocka R.}:
\newblock 3d highlighter: Localizing regions on 3d shapes via text descriptions.
\newblock \emph{arXiv preprint arXiv:2212.11263} (2022).

\bibitem[DXXD22]{ding2022decoupling}
\textsc{Ding J., Xue N., Xia G.-S., Dai D.}:
\newblock {Decoupling Zero-Shot Semantic Segmentation}.
\newblock In \emph{Proceedings of the IEEE/CVF Conference on Computer Vision and Pattern Recognition (CVPR)} (2022), pp.~11583--11592.

\bibitem[FWJ{\etalchar{*}}22]{fan2022nerf}
\textsc{Fan Z., Wang P., Jiang Y., Gong X., Xu D., Wang Z.}:
\newblock Nerf-sos: Any-view self-supervised object segmentation on complex scenes.
\newblock \emph{arXiv preprint arXiv:2209.08776} (2022).

\bibitem[FZC{\etalchar{*}}22]{fu2022panoptic}
\textsc{Fu X., Zhang S., Chen T., Lu Y., Zhu L., Zhou X., Geiger A., Liao Y.}:
\newblock Panoptic nerf: 3d-to-2d label transfer for panoptic urban scene segmentation.
\newblock In \emph{International Conference on 3D Vision (3DV)} (2022).

\bibitem[GGCL22]{ghiasi2022scaling}
\textsc{Ghiasi G., Gu X., Cui Y., Lin T.-Y.}:
\newblock {Scaling Open-Vocabulary Image Segmentation with Image-Level Labels}.
\newblock In \emph{Proceedings of the European Conference on Computer Vision (ECCV)} (2022), pp.~540--557.

\bibitem[HCJW22]{huang2022multi}
\textsc{Huang S., Chen Y., Jia J., Wang L.}:
\newblock Multi-view transformer for 3d visual grounding.
\newblock In \emph{CVPR} (2022).

\bibitem[IMK20]{iqbal2020weakly}
\textsc{Iqbal U., Molchanov P., Kautz J.}:
\newblock Weakly-supervised 3d human pose learning via multi-view images in the wild.
\newblock In \emph{Proceedings of the IEEE/CVF conference on computer vision and pattern recognition} (2020), pp.~5243--5252.

\bibitem[JYX{\etalchar{*}}21]{jia2021scaling}
\textsc{Jia C., Yang Y., Xia Y., Chen Y.-T., Parekh Z., Pham H., Le Q., Sung Y.-H., Li Z., Duerig T.}:
\newblock {Scaling Up Visual and Vision-Language Representation Learning with Noisy Text Supervision}.
\newblock In \emph{Proceedings of the International Conference on Machine Learning (ICML)} (2021), pp.~4904--4916.

\bibitem[KGY{\etalchar{*}}22]{kundu2022panoptic}
\textsc{Kundu A., Genova K., Yin X., Fathi A., Pantofaru C., Guibas L.~J., Tagliasacchi A., Dellaert F., Funkhouser T.}:
\newblock {Panoptic Neural Fields: A Semantic Object-Aware Neural Scene Representation}.
\newblock In \emph{Proceedings of the IEEE/CVF Conference on Computer Vision and Pattern Recognition (CVPR)} (2022), pp.~12871--12881.

\bibitem[KKG{\etalchar{*}}23]{kerr2023lerf}
\textsc{Kerr J., Kim C.~M., Goldberg K., Kanazawa A., Tancik M.}:
\newblock Lerf: Language embedded radiance fields.
\newblock In \emph{Proceedings of the IEEE/CVF International Conference on Computer Vision (ICCV)} (October 2023), pp.~19729--19739.

\bibitem[KMS22]{kobayashi2022decomposing}
\textsc{Kobayashi S., Matsumoto E., Sitzmann V.}:
\newblock {Decomposing NeRF for Editing via Feature Field Distillation}.
\newblock In \emph{Advances in Neural Information Processing Systems (NeurIPS)} (2022).

\bibitem[LE22]{luddecke2022image}
\textsc{L{\"u}ddecke T., Ecker A.}:
\newblock {Image Segmentation Using Text and Image Prompts}.
\newblock In \emph{Proceedings of the IEEE/CVF Conference on Computer Vision and Pattern Recognition (CVPR)} (2022), pp.~7086--7096.

\bibitem[LS18]{li2018megadepth}
\textsc{Li Z., Snavely N.}:
\newblock Megadepth: Learning single-view depth prediction from internet photos.
\newblock In \emph{Proceedings of the IEEE conference on computer vision and pattern recognition} (2018), pp.~2041--2050.

\bibitem[LWB{\etalchar{*}}22]{li2022language}
\textsc{Li B., Weinberger K.~Q., Belongie S., Koltun V., Ranftl R.}:
\newblock {Language-Driven Semantic Segmentation}.
\newblock In \emph{Proceedings of the International Conference on Learning Representations (ICLR)} (2022).

\bibitem[LWD{\etalchar{*}}23]{liang2023open}
\textsc{Liang F., Wu B., Dai X., Li K., Zhao Y., Zhang H., Zhang P., Vajda P., Marculescu D.}:
\newblock Open-vocabulary semantic segmentation with mask-adapted clip.
\newblock In \emph{Proceedings of the IEEE/CVF Conference on Computer Vision and Pattern Recognition (CVPR)} (June 2023), pp.~7061--7070.

\bibitem[LXW{\etalchar{*}}23]{lu2023open}
\textsc{Lu Y., Xu C., Wei X., Xie X., Tomizuka M., Keutzer K., Zhang S.}:
\newblock Open-vocabulary point-cloud object detection without 3d annotation.
\newblock In \emph{Proceedings of the IEEE/CVF Conference on Computer Vision and Pattern Recognition (CVPR)} (June 2023), pp.~1190--1199.

\bibitem[MBRS{\etalchar{*}}21]{martin2021nerf}
\textsc{Martin-Brualla R., Radwan N., Sajjadi M.~S., Barron J.~T., Dosovitskiy A., Duckworth D.}:
\newblock Nerf in the wild: Neural radiance fields for unconstrained photo collections.
\newblock In \emph{Proceedings of the IEEE/CVF Conference on Computer Vision and Pattern Recognition} (2021), pp.~7210--7219.

\bibitem[MST{\etalchar{*}}20]{mildenhall2020nerf}
\textsc{Mildenhall B., Srinivasan P.~P., Tancik M., Barron J.~T., Ramamoorthi R., Ng R.}:
\newblock {NeRF: Representing Scenes as Neural Radiance Fields for View Synthesis}.
\newblock In \emph{Proceedings of the European Conference on Computer Vision (ECCV)} (2020), pp.~405--421.

\bibitem[PGJ{\etalchar{*}}22]{peng2022openscene}
\textsc{Peng S., Genova K., Jiang C., Tagliasacchi A., Pollefeys M., Funkhouser T., et~al.}:
\newblock {OpenScene: 3D Scene Understanding with Open Vocabularies}.
\newblock \emph{arXiv preprint arXiv:2211.15654} (2022).

\bibitem[RKH{\etalchar{*}}21]{radford2021learning}
\textsc{Radford A., Kim J.~W., Hallacy C., Ramesh A., Goh G., Agarwal S., Sastry G., Askell A., Mishkin P., Clark J., Krueger G., Sutskever I.}:
\newblock {Learning Transferable Visual Models from Natural Language Supervision}.
\newblock In \emph{Proceedings of the International Conference on Machine Learning (ICML)} (2021), pp.~8748--8763.

\bibitem[RLD22]{rozenberszki2022language}
\textsc{Rozenberszki D., Litany O., Dai A.}:
\newblock {Language-Grounded Indoor 3D Semantic Segmentation in the Wild}.
\newblock In \emph{Proceedings of the European Conference on Computer Vision (ECCV)} (2022), pp.~125--141.

\bibitem[RMBB{\etalchar{*}}13]{russell20133d}
\textsc{Russell B.~C., Martin-Brualla R., Butler D.~J., Seitz S.~M., Zettlemoyer L.}:
\newblock 3d wikipedia: Using online text to automatically label and navigate reconstructed geometry.
\newblock \emph{ACM Transactions on Graphics (TOG) 32}, 6 (2013), 1--10.

\bibitem[SF16]{schonberger2016structure}
\textsc{Schonberger J.~L., Frahm J.-M.}:
\newblock Structure-from-motion revisited.
\newblock In \emph{Proceedings of the IEEE conference on computer vision and pattern recognition} (2016), pp.~4104--4113.

\bibitem[SGSS08]{snavely2008finding}
\textsc{Snavely N., Garg R., Seitz S.~M., Szeliski R.}:
\newblock Finding paths through the world's photos.
\newblock \emph{ACM Transactions on Graphics (TOG) 27}, 3 (2008), 1--11.

\bibitem[SPB{\etalchar{*}}22]{siddiqui2022panoptic}
\textsc{Siddiqui Y., Porzi L., Bul{\'o} S.~R., M{\"u}ller N., Nie{\ss}ner M., Dai A., Kontschieder P.}:
\newblock {Panoptic Lifting for 3D Scene Understanding with Neural Fields}.
\newblock \emph{arXiv preprint arXiv:2212.09802} (2022).

\bibitem[SSS06]{snavely2006photo}
\textsc{Snavely N., Seitz S.~M., Szeliski R.}:
\newblock Photo tourism: exploring photo collections in 3d.
\newblock In \emph{ACM siggraph 2006 papers} (2006), pp.~835--846.

\bibitem[SSW{\etalchar{*}}21]{sun2021loftr}
\textsc{Sun J., Shen Z., Wang Y., Bao H., Zhou X.}:
\newblock Loftr: Detector-free local feature matching with transformers.
\newblock In \emph{Proceedings of the IEEE/CVF conference on computer vision and pattern recognition} (2021), pp.~8922--8931.

\bibitem[TLLV22]{tschernezki22neural}
\textsc{Tschernezki V., Laina I., Larlus D., Vedaldi A.}:
\newblock {Neural Feature Fusion Fields: 3D Distillation of Self-Supervised 2D Image Representations}.
\newblock In \emph{Proceedings of the International Conference on 3D Vision (3DV)} (2022).

\bibitem[TWN{\etalchar{*}}23]{tancik2023nerfstudio}
\textsc{Tancik M., Weber E., Ng E., Li R., Yi B., Kerr J., Wang T., Kristoffersen A., Austin J., Salahi K., et~al.}:
\newblock Nerfstudio: A modular framework for neural radiance field development.
\newblock \emph{arXiv preprint arXiv:2302.04264} (2023).

\bibitem[TZFR23]{turki2023suds}
\textsc{Turki H., Zhang J.~Y., Ferroni F., Ramanan D.}:
\newblock Suds: Scalable urban dynamic scenes.
\newblock In \emph{Proceedings of the IEEE/CVF Conference on Computer Vision and Pattern Recognition (CVPR)} (2023), pp.~12375--12385.

\bibitem[WAESS21]{wu2021towers}
\textsc{Wu X., Averbuch-Elor H., Sun J., Snavely N.}:
\newblock {Towers of Babel: Combining Images, Language, and 3D Geometry for Learning Multimodal Vision}.
\newblock In \emph{Proceedings of the IEEE/CVF International Conference on Computer Vision (ICCV)} (2021), pp.~428--437.

\bibitem[WZHS20]{wang2020learning}
\textsc{Wang Q., Zhou X., Hariharan B., Snavely N.}:
\newblock Learning feature descriptors using camera pose supervision.
\newblock In \emph{Computer Vision--ECCV 2020: 16th European Conference, Glasgow, UK, August 23--28, 2020, Proceedings, Part I 16} (2020), Springer, pp.~757--774.

\bibitem[XDML{\etalchar{*}}22]{xu2022groupvit}
\textsc{Xu J., De~Mello S., Liu S., Byeon W., Breuel T., Kautz J., Wang X.}:
\newblock {GroupViT: Semantic Segmentation Emerges from Text Supervision}.
\newblock In \emph{Proceedings of the IEEE/CVF Conference on Computer Vision and Pattern Recognition (CVPR)} (2022), pp.~18134--18144.

\bibitem[XXP{\etalchar{*}}21]{xiangli2021citynerf}
\textsc{Xiangli Y., Xu L., Pan X., Zhao N., Rao A., Theobalt C., Dai B., Lin D.}:
\newblock Citynerf: Building nerf at city scale.
\newblock \emph{arXiv preprint arXiv:2112.05504} (2021).

\bibitem[XZW{\etalchar{*}}21]{xu2021simple}
\textsc{Xu M., Zhang Z., Wei F., Lin Y., Cao Y., Hu H., Bai X.}:
\newblock {A Simple Baseline for Open-Vocabulary Semantic Segmentation with Pre-trained Vision-language Model}.
\newblock \emph{arXiv preprint arXiv:2112.14757} (2021).

\bibitem[Yi20]{yi2020image}
\textsc{Yi K.~M.}:
\newblock Image matching: Local features \& beyond 2020.
\newblock https://www.cs.ubc.ca/~kmyi/imw2020/data.html, 2020.
\newblock https://www.cs.ubc.ca/~kmyi/imw2020/data.html.
\newblock URL: \url{https://www.cs.ubc.ca/~kmyi/imw2020/data.html}, \href {http://arxiv.org/abs/https://www.cs.ubc.ca/~kmyi/imw2020/data.html} {\path{arXiv:https://www.cs.ubc.ca/~kmyi/imw2020/data.html}}.

\bibitem[ZLD22]{zhou2022maskclip}
\textsc{Zhou C., Loy C.~C., Dai B.}:
\newblock {Extract Free Dense Labels from CLIP}.
\newblock In \emph{Proceedings of the European Conference on Computer Vision (ECCV)} (2022), pp.~696--712.

\bibitem[ZLLD21]{zhi2021in-place}
\textsc{Zhi S., Laidlow T., Leutenegger S., Davison A.~J.}:
\newblock {In-Place Scene Labelling and Understanding with Implicit Scene Representation}.
\newblock In \emph{Proceedings of the IEEE/CVF International Conference on Computer Vision (CVPR)} (2021), pp.~15838--15847.

\bibitem[ZZP{\etalchar{*}}16]{zhou2016semantic}
\textsc{Zhou B., Zhao H., Puig X., Fidler S., Barriuso A., Torralba A.}:
\newblock Semantic understanding of scenes through the ade20k dataset.
\newblock \emph{arXiv preprint arXiv:1608.05442} (2016).

\bibitem[ZZP{\etalchar{*}}17]{zhou2017scene}
\textsc{Zhou B., Zhao H., Puig X., Fidler S., Barriuso A., Torralba A.}:
\newblock Scene parsing through ade20k dataset.
\newblock In \emph{Proceedings of the IEEE Conference on Computer Vision and Pattern Recognition} (2017).

\end{thebibliography}

\clearpage
\medskip \noindent \textbf{Supplementary Material
}

\title{Supplementary Material for \ourmethod{} }
\maketitle

\setcounter{section}{0}
\section{\ourdataset{} -- Additional Details}

\begin{figure*}
\centering
\includegraphics[width=\linewidth]{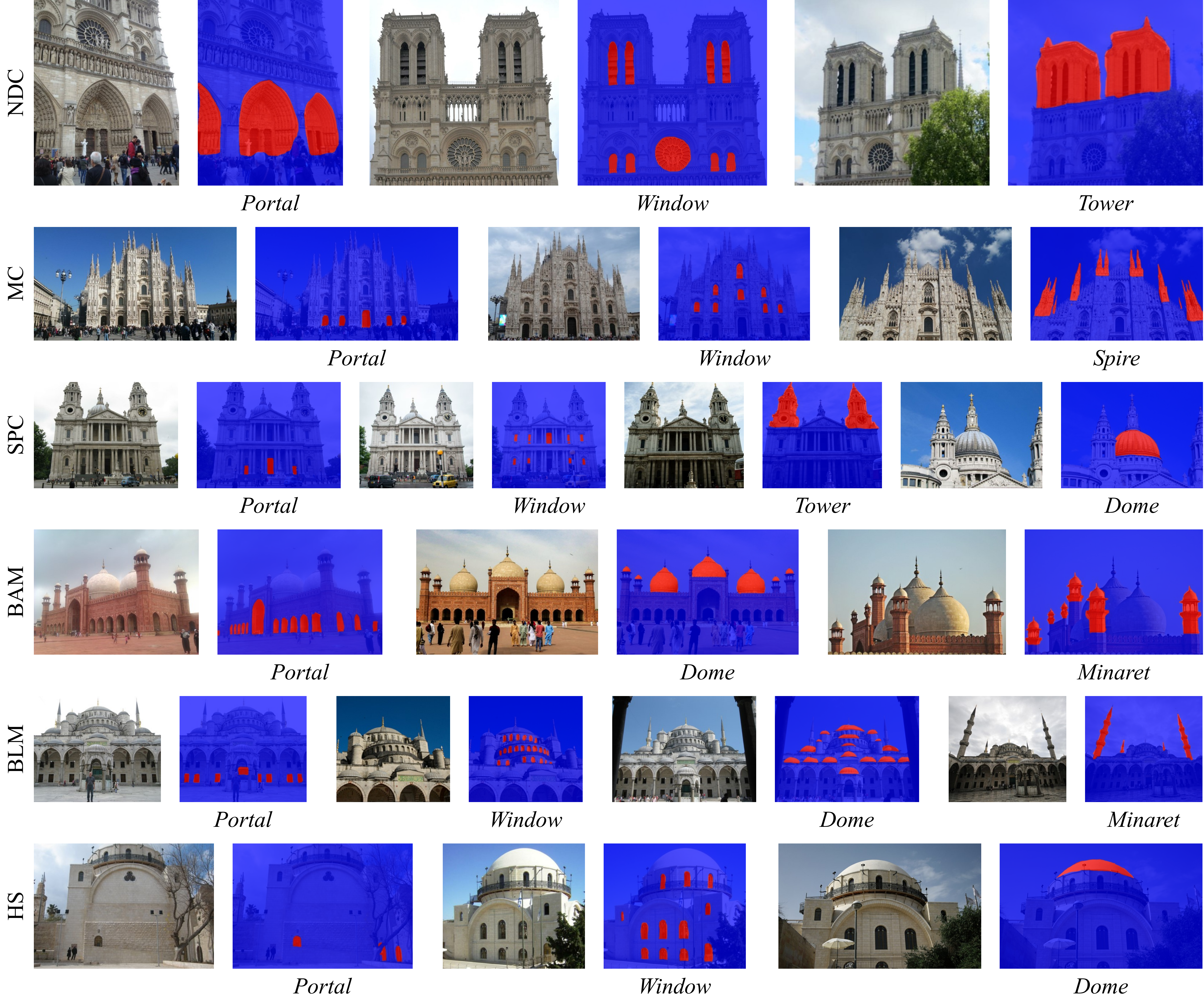}
\caption{\textbf{\ourdataset{} annotations.} We illustrate annotations for each category in each landmark in the \ourdataset{} dataset: Notre-Dame Cathedral (NDC), Milan Cathedral (MC), St. Paul's Cathedral (SPC), Badshahi Mosque (BAM), Blue Mosque (BLM), and Hurva Synagogue (HS).}
\label{fig:dataset}
\end{figure*}

\medskip \noindent \textbf{Landmarks and Categories Used}.

Our benchmark spans three landmark building types (cathedrals, mosques, and a synagogue), from different areas around the world. We select scenes that have sufficient RGB imagery for reconstructing with \cite{chen2022hallucinated}. The images were taken from IMC-PT 20~\cite{yi2020image} (\emph{Notre-Dame
Cathedral}, \emph{St. Paul’s Cathedral}), MegaDepth~\cite{li2018megadepth} (\emph{Blue Mosque}), WikiScenes~\cite{wu2021towers} (\emph{Milan Cathedral}), and scraped from Wikimedia Commons using the WikiScenes data scraping procedure (\emph{Badshahi Mosque} and \emph{Hurva Synagogue}).  The Notre-Dame cathedral has the most images in the dataset (3,765 images), and the \emph{Hurva Synagogue} has the fewest (104 images). For semantic categories, we select diverse concepts of different scales. Some of these (such as \emph{portal}) are applicable to all landmarks in our dataset while others (such as \emph{minaret}) only apply to certain landmarks. As illustrated in Table \ref{tab:dataset}, we provide segmentations of 3-4 semantic categories for each landmark; these are selected based on the relevant categories in each case (\emph{e.g.} only the two mosques have minarets).

\medskip \noindent \textbf{Annotation Procedure}

We produce ground-truth binary segmentation maps to evaluate our method using manual labelling combined with correspondence-guided propagation. We first segment 110 images from 3-4 different categories from each of the six different scenes in our dataset, as shown in Table \ref{tab:dataset}. We then estimate homographies between these images and the remaining images for these landmarks, using shared keypoint correspondences from COLMAP~\cite{schonberger2016structure} and RANSAC. We require at least 100 corresponding keypoints that are RANSAC inliers; we also filter out extreme (highly skewed or rotated) homographies by using the condition number of the first two columns of the homography matrix. %
When multiple propagated masks can be inferred for a target image, we calculate each pixel's binary value by a majority vote of the warped masks. Finally, we filter these augmented masks by manual inspection. Out of 8,951 images, 6,195 were kept (along with the original manual seeds), resulting in a final benchmark size of 6,305 items.  Those that were filtered are mostly due to occlusions and inaccurate warps. Annotation examples from our benchmark are shown in Figure \ref{fig:dataset}.

\begin{table}[!t]
\centering
\resizebox{\linewidth}{!}{
\begin{tabular}{lcccccccc}
\toprule

Landmark  & Portal&  Window& Spire & Tower & Dome & Minaret &  \#Seed & \#Seg\\ 
\midrule
NDC & $\cmark$ & $\cmark$ & $\xmark$ & $\cmark$ & $\xmark$ & $\xmark$ & 15 & 4048  \\ 
MC & $\cmark$ & $\cmark$ & $\cmark$ & $\xmark$ & $\xmark$ & $\xmark$& 19 & 902    \\ 
SPC & $\cmark$	& $\cmark$	& $\xmark$	& $\cmark$ & $\cmark$  &  $\xmark$ & 20 & 731     \\ 
BAM & $\cmark$	&  $\xmark$	& $\xmark$	& $\xmark$ & $\cmark$ & $\cmark$ & 15 & 177   \\ 
BLM & $\cmark$ & $\cmark$	&  $\xmark$ & $\xmark$ & $\cmark$ & $\cmark$ & 26    & 381 \\ 
HS & $\cmark$	& $\cmark$	&  $\xmark$ &  $\xmark$ &$\cmark$  &  $\xmark$ & 15 & 66    \\ 

\bottomrule

\end{tabular}
}
\caption{\textbf{The \ourdataset{} Benchmark}, composed of the Notre-Dame Cathedral (NDC), Milan Cathedral (MC), St. Paul's Cathedral (SPC), Badshahi Mosque (BAM), Blue Mosque (BLM), and the Hurva Synagogue (HS). Above we report the set of semantic categories annotated for each landmark, chosen according to their visible structure. In the columns on the right, we report the number of initial manually segmented images (\#Seed), and the final number of ground-truth segmentations after augmented with filtered warps (\#Seg)
}
\label{tab:dataset}
\end{table}

\def\thefootnote{*}\footnotetext{Denotes equal contribution}

\section{Implementation Details}

\subsection{Augmenting the WikiScenes Dataset}

The original WikiScenes dataset is as described in Wu \etal~\shortcite{wu2021towers}. To produce training data for the offline stages of our system (LLM-based semantic distillation and V\&L model semantic adaptation), we augment this cathedral-focused dataset with mosques by using the same procedure to scrape freely-available Wikimedia Commons collected from the root WikiCategory ``Mosques by year of completion''. The collected data contains a number of duplicate samples, since the same image may appear under different categories in Wikimedia Commons and is thus retrieved multiple times by the scraping script. In order to de-duplicate, we treat the image's filename (as accessed on Wikimedia Commons) as a unique identifier. After de-duplication, we are left with 69,085 cathedral images and 45,668 mosque images. Out of these, we set aside the images from landmarks which occur in \ourdataset{} (13,743 images total) to prevent test set leakage; the remaining images serve as our training data.

\subsection{LLM-Based Semantic Distillation}

To distill the image metadata into concise textual pseudo-labels, we use the instruction-tuned language model Flan-T5~\cite{chung2022scaling}, selecting the 3B parameter Flan-T5-XL variant. The model is given the image caption, related key-words, and filename, and outputs a single word describing a prominent architectural feature within the image serving as its pseudo-label. Text is generated using beam search decoding with four beams. The prompt given to Flan-T5 includes the instruction to \emph{Write ``unknown'' if it is not specified} (i.e. the architectural feature), in order to allow the language model to express uncertainty instead of hallucinating incorrect answers in indeterminate cases, as described in our main paper. We also find the use of the building's name in the prompt (\emph{What architectural feature of \B...}) to be important in order to cue the model to omit the building's name from its output (e.g. \emph{towers of the Cathedral of Seville} vs. simply \emph{towers}).

To post-process these labels, we employ the following textual cleanup techniques. We (1) employ lowercasing, (2) remove outputs starting with ``un-'' (``unknown'', ``undefined'' etc.), and (3) remove navigation words (e.g. ``west'' in ``west facade'') since these are not informative for learning visual semantics. Statistics on the final pseudo-labels are given in Section \ref{sec:supp_pl_stats}.

\subsection{Semantic Adaptation of V\&L Models}

We fine-tune \ourclip{} on images and associated pseudo-labels, preprocessing by removing all pairs whose pseudo-label begins with ``un-'' (e.g. ``unknown'', ``undetermined'', etc.) and removing initial direction words (``north'', ``southern'', ``north eastern'', etc.), as these are not visually informative. In total, this consists of 57,874 such pairs used as training data representing 4,031 unique pseudo-label values; this includes 41,452 pairs from cathedrals and 16,422 pairs from mosques. Fine-tuning is performed on CLIP initialized with the \texttt{clip-ViT-B-32} checkpoint as available in the \texttt{sentence-transformers} model collection on Hugging Face model hub, using the contrastive multiple negatives ranking loss as implemented in the Sentence Transformers library. We train for 5 epochs with learning rate 1e-6 and batch size 128.

To collect training data based on image correspondences for \ourclipseg{}, we use the following procedure: Firstly, to find pairs of images in geometric correspondence, we perform a search on pairs of images $(I_1, I_2)$ from each building in our train set along with the pseudo-label $P$ of the first image, applying LoFTR~\cite{sun2021loftr} to such pairs and filtering for pairs in correspondence where $I_1$ is a zoomed-in image corresponding to category $P$ and $I_2$ is a corresponding zoomed-out image. We filter correspondences using the following heuristic requirements:
\begin{itemize}
    \item At least 50 corresponding keypoints that are inliers using OpenCV's \texttt{USAC\_MAGSAC} method for fundamental matrix calculation.
    \item A log-ratio of at least 0.1 between the dispersion (mean square distance from centroid, using relative distances to the image dimensions) of inlier keypoints in $I_1$ and $I_1$.
    \item \ourclip{} similarity of at least 0.2 between $I_1$ and $P$, and at most 0.3 between $I_2$ and $P$. This is because $I_1$ should match $P$, while $I_2$, as a zoomed-out image, should contain $P$ but not perfectly match it as a concept.
    \item At least 3 inlier keypoints within the region $R_P$ of $I_1$ matching $P$. $R_P$ is estimated by by segmenting $I_1$ with CLIPSeg and prompt $P$ and binarizing with threshold $0.3$.
    \item A low ratio of areas of the region matching $P$ relative to the building's facade, since this suggests a localizable concept. This is estimated as follows: We first find the quadrilateral Q which is the region of $I_2$ corresponding to $I_1$, by projecting $I_1$ with the homography estimated from corresponding keypoints. We then find the facade of the building in $I_2$ by segmenting $I_2$ using CLIPSeg with the prompt \emph{cathedral} or \emph{mosque} (as appropriate for the given landmark), which outputs the matrix of probabilities $M$. Finally, we calculate the sum of elements of $M$ contained in $Q$ divided by the sum of all elements of $M$, and check if this is less than $0.5$.
\end{itemize} 

Empirically, we find that these heuristics succeed in filtering out many types of uninteresting image pairs and noise while selecting for the correspondences and pseudo-labels that are of interest. Due to computational constraints, we limit our search to 50 images from each landmark in our train set paired with every other image from the same landmark, and this procedure yields 3,651 triplets $(I_1, I_2, P)$ in total, covering 181 unique pseudo-label categories. To use these correspondences as supervision for training segmentation, we segment $I_1$ using CLIPSeg with prompt $P$, project this segmentation onto $I_2$ using the estimated homography, and using the resulting segmentation map in the projected region as ground-truth for segmenting $I_2$ with $P$.

In addition to this data, we collect training data on a larger scale by searching for images from the entire training dataset with crops that are close to particular pseudo-labels. To do this, we run a search by randomly selecting landmark $L$ and and one of its images $I$, selecting a random pseudo-label $P$ that appears with $L$ (not necessarily with the chosen image) in our dataset, selecting a random crop $C$ of $I$, and checking its similarity to $P$ with \ourclip{}. We check if the following heuristic conditions hold:

\begin{itemize}
    \item $C$ must have \ourclip{} similarity of at least 0.2 with $P$.
    \item $C$ must have higher \ourclip{} similarity to $P$ than $I$ does.
    \item This similarity must be higher than the similarity between $C$ and the 20 most common pseudo-labels in our train dataset (excluding $P$, if it is one of these common pseudo-labels).
    \item $C$ when segmented using CLIPSeg with prompt $P$ must have some output probability at least $0.1$ in its central area (the central 280$\times$280 region within the 352$\times$352 output matrix).
\end{itemize}

If these conditions hold, we use the pair $(I, P)$ along with the CLIPSeg segmentation of the crop $C$ with prompt $P$ as ground-truth data for fine-tuning our segmentation model. Although this search could be run indefinitely, we terminate it after collecting 29,440 items to use as training data.

For both sources of data (correspondence-based and crop-based), we further refine the pseudo-labels by converting them to singular, removing digits and additional direction words, and removing non-localizable concepts and those referring to most of the landmark or its entirety (``mosque'', ``front'', ``gothic'', ``cathedral'', ``side'', ``view'').

We fine-tune CLIPSeg to produce \ourclipseg{} by training for 10 epochs with learning rate 1e-4. We freeze CLIPSeg's encoders and only train its decoder module. To provide robustness to label format, we randomly augment textual pseudo-labels by converting them from singular to plural form (e.g. ``window'' $\to$ ``windows'') with probability $0.5$. At each iteration, we calculate losses using a single image and ground-truth pair from the correspondence-based data, and a minibatch of four image and ground-truth pairs from the crop-based data. We use four losses for training, summed together with equal weighting, as described in the main paper in Section 3.2.

CLIPSeg (and \ourclipseg{}) requires a square input tensor with spatial dimensions 352$\times$352. In order to handle images of varying aspect ratios during inference, we apply vertical replication padding to short images, and to wide images we average predictions applied to a horizontally sliding window. In the latter case, we use overlapping windows with stride of 25 pixels, after resizing images to have maximum dimension of size 500 pixels. Additionally, in outdoor scenes, we apply inference after zooming in to the bounding box of the building in question, in order to avoid attending to irrelevant regions. The building is localized by applying CLIPSeg with the zero-shot prompt \emph{cathedral}, \emph{mosque}, or \emph{synagogue} (as appropriate for the building in question), selecting the smallest bounding box containing all pixels with predicted probabilities above $0.5$, and adding an additional 10\% margin on all sides. While our model may accept arbitrary text as input, we normalize inputs for metric calculations to plural form (``portals'', ``windows'', ``spires'' etc.) for consistency.

\subsection{3D Localization}

We build on top of the Ha-NeRF~\cite{chen2022hallucinated} architecture with an added semantic channel, similarly to Zhi \etal~\cite{zhi2021in-place}. This semantic channel consists of an MLP with three hidden layers (dimensions 256, 256, 128) with ReLU activations, and a final output layer for binary prediction with a softmax activation. We first train the Ha-NeRF RGB model of a scene (learning rate 5e-4 for 250K iterations); we then freeze the shared MLP backbone of the RGB and semantic channels and train only the semantic channel head (learning rate 5e-5, 12.5K iterations). We train with batch size 8,192. When training the semantic channel, the targets are binary segmentation masks produced by \ourclipseg{} with a given text prompt, using the inference method described above. We binarize these targets (threshold 0.2) to reduce variance stemming from outputs with low confidence, and we use a binary cross-entropy loss function when training on them.

For indoor scenes, we use all available images to train our model. For outdoor scenes, we select 150 views with segmentations for building the 3D semantic field by selecting for images with clear views of the building's entire facade without occlusions. We find that this procedure yields comparable performance to using all the images in the collection, while being more computationally efficient. To select these images, we first segment each candidate image with CLIPSeg using one of the prompts \emph{cathedral}, \emph{mosque}, or \emph{synagogue} (as relevant) , select the largest connected component $C$ of the output binary mask (using thresold 0.5), and sort the images by the minimum horizontal or vertical margin length of this component from the image's borders. This prioritizes images where the building facade is fully visible and contained within the boundary of the visible image. To prevent occluded views of the building from being selected, we add a penalty using the proportion of overlap $C$ and the similar binary mask $C'$ calculated on the RGB NeRF reconstruction of the same view, since transient occlusions are not typically reconstructed by the RGB NeRF. In addition, we penalize images with less than 10\% or more than 90\% total area covered by $C$, since these often represent edge cases where the building is barely visible or not fully contained within the image. Written precisely, the scoring formula is given by $s = m + c - x$, where $m$ is the aforementioned margin size (on a scale from $0$ to $1$), $c$ is the proportion of area of $C'$ overlapping $C$, and $x$ is a penalty of $1.0$ when $C$ covers too little or much of the image (as described before) and $0$ otherwise.

\medskip \noindent \textbf{Runtime}. A typical run (optimizing the volumetric probabilities for a single landmark) takes roughly 2 hours on a NVIDIA RTX A5000 with a single GPU. Optimizing the RGB and density values is only done once per landmark, and takes 2 days on average, depending on the number of images in the collection.

\subsection{Baseline Comparisons} \label{sec:comparison_details}

We provide additional details of our comparison to DFF~\cite{kobayashi2022decomposing} and LERF~\cite{kerr2023lerf}. We train these models on the same images used to train our model. We use a Ha-NeRF backbone; similarly to our method we train the RGB NeRF representations for 250K steps and then the semantic representations for an additional 150K steps. Otherwise follow the original training and implementation details of these models, which we reproduce here for clarity.

For DFF, we implement feature layers as an MLP with 2 hidden layers of 128 and ReLU activations. The input to the DFF model is the images and the corresponding features derived from LSeg, and we minimize the difference between the learned features and LSeg features with an L2 loss, training with batch size 1024.

For LERF, we use the official implementation which uses the Nerfacto method and 
 the Nerfstudio API~\cite{tancik2023nerfstudio}. The architecture includes a DINO MLP with one hidden layer of dimension 256 and a ReLU activation; and a CLIP MLP consisting of with 3 hidden layers of dimension 256, ReLU activations, and a final 512-dimensional output layer. The input to this model consists of images, their CLIP embeddings in different scales, and their DINO features. We use the same loss as the original LERF paper: CLIP loss for the CLIP embeddings to maximize the cosine similarity, and MSE loss for the DINO features. The CLIP loss is multiplied by a factor of 0.01 similar to the LERF paper. We use an image pyramid from scale 0.05 to 0.5 in 7 steps. We train this model with batch size was 4096. We used also the relevancy score with the same canonical phrases as described in the LERF paper: ``object'', ``things'', ``stuff'', and ``texture''.

\section{Additional Results and Ablations}
\begin{figure*}
\centering
    \rotatebox{90}{\;\;\;\;\;\ourclipseg{}}
    \jsubfig{\includegraphics[height=2.33cm]{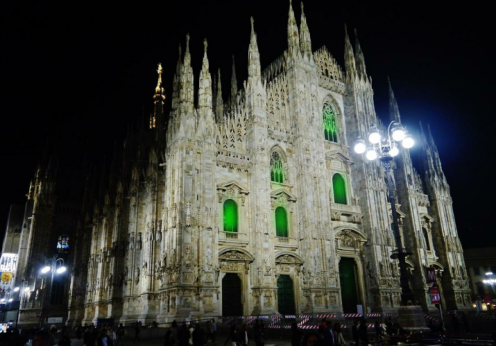}}{\textit{}}
    \jsubfig{\includegraphics[height=2.33cm]{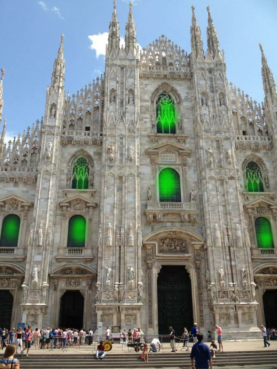}}{\textit{}}
    \jsubfig{\includegraphics[height=2.33cm]{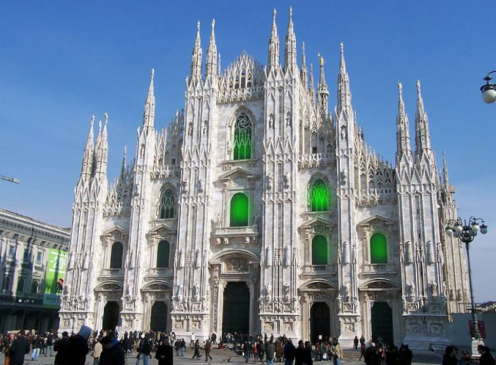}}{\textit{}}
    \hfill
    \rotatebox{90}{\;\;\ourclipseg{} (th)}
    \jsubfig{\includegraphics[height=2.33cm]{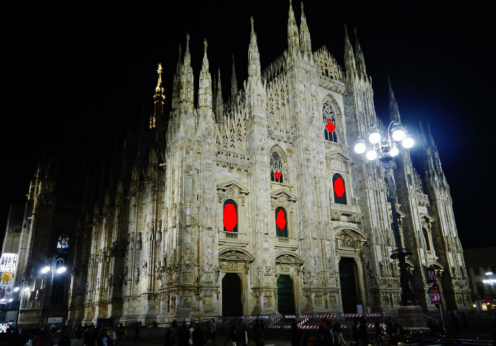}}{\textit{}}
    \jsubfig{\includegraphics[height=2.33cm]{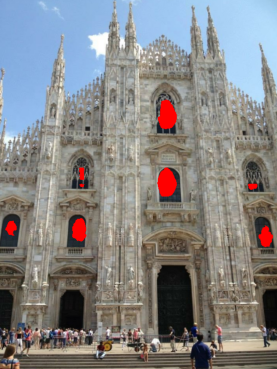}}{\textit{}}
    \jsubfig{\includegraphics[height=2.33cm]{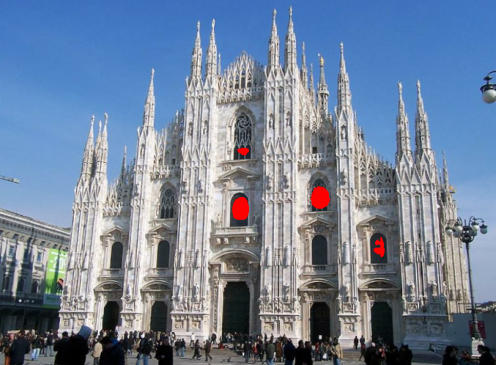}}{\textit{}}
    
    \rotatebox{90}{\;\;\;\;\,\ourmethod{}}
    \jsubfig{\includegraphics[height=2.33cm]{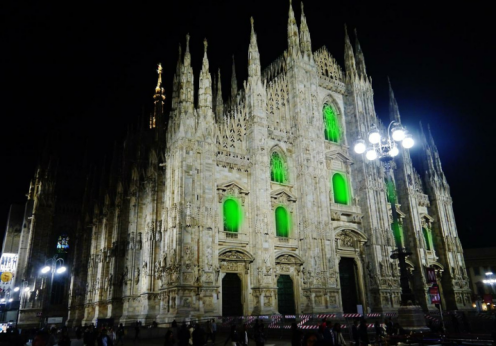}}{\textit{}}
    \jsubfig{\includegraphics[height=2.33cm]{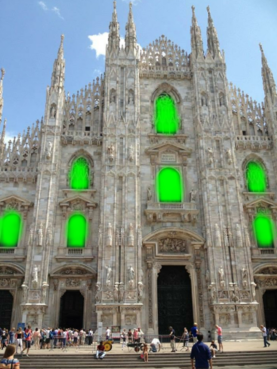}}{\textit{}}
    \jsubfig{\includegraphics[height=2.33cm]{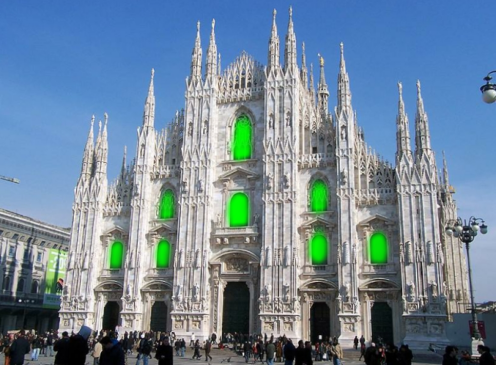}}{\textit{}}
    \hfill
    \rotatebox{90}{\;\,\ourmethod{} (th)}
    \jsubfig{\includegraphics[height=2.33cm]{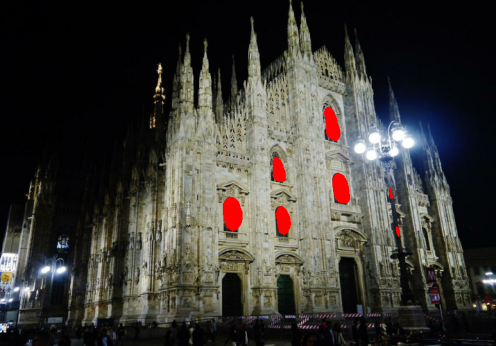}}{\textit{}}
    \jsubfig{\includegraphics[height=2.33cm]{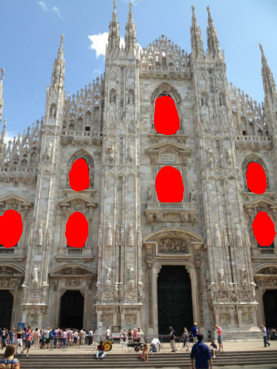}}{\textit{}}
    \jsubfig{\includegraphics[height=2.33cm]{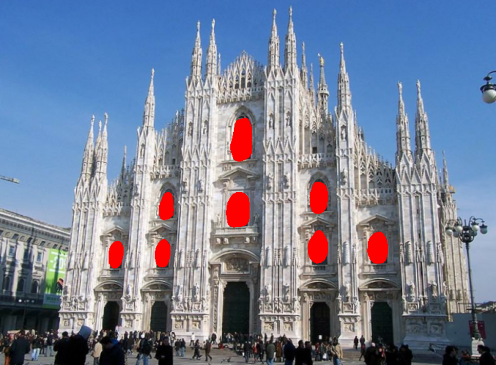}}{\textit{}}

    \rotatebox{90}{\;\;\;\;\;\;\ourclipseg{}}
    \jsubfig{\includegraphics[height=2.85cm]{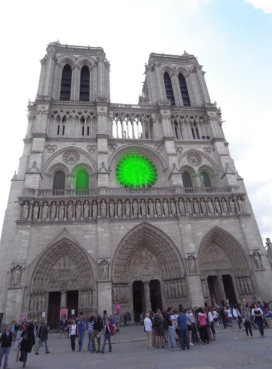}}{\textit{}}
    \jsubfig{\includegraphics[height=2.85cm]{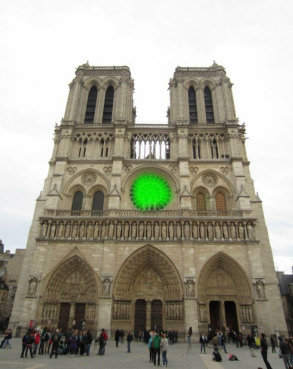}}{\textit{}}
    \jsubfig{\includegraphics[height=2.85cm]{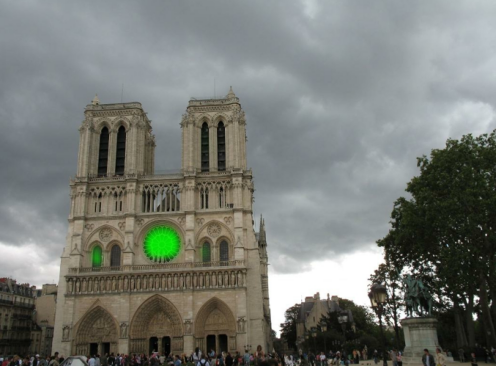}}{\textit{}}
    \hfill
    \rotatebox{90}{\;\;\;\;\ourclipseg{} (th)}
     \jsubfig{\includegraphics[height=2.85cm]{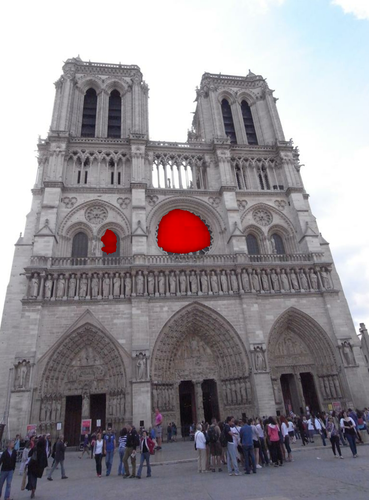}}{\textit{}}
    \jsubfig{\includegraphics[height=2.85cm]{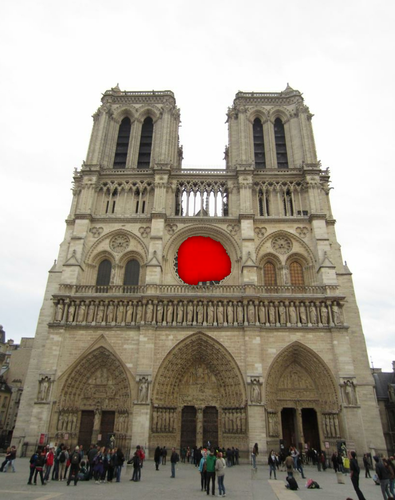}}{\textit{}}
    \jsubfig{\includegraphics[height=2.85cm]{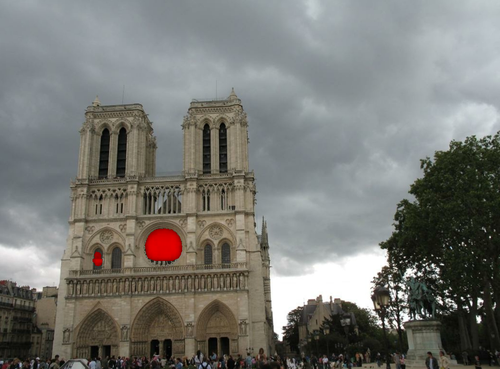}}{\textit{}}
    
    \rotatebox{90}{\;\;\;\;\;\,\ourmethod{}}
    \jsubfig{\includegraphics[height=2.85cm]{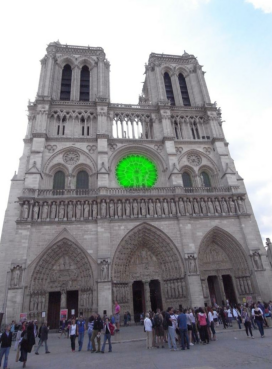}}{\textit{}}
    \jsubfig{\includegraphics[height=2.85cm]{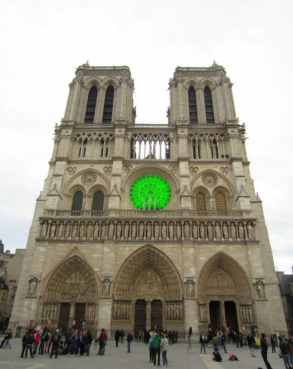}}{\textit{}}
    \jsubfig{\includegraphics[height=2.85cm]{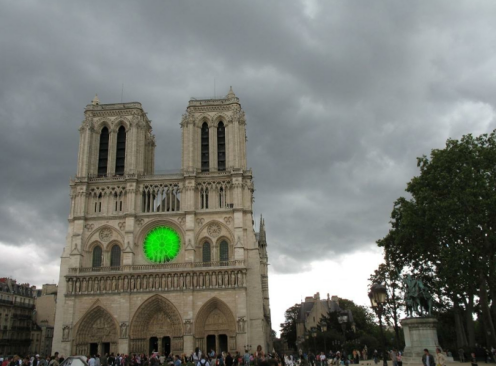}}{\textit{}}
    \hfill
    \rotatebox{90}{\;\;\;\,\ourmethod{} (th)}
    \jsubfig{\includegraphics[height=2.85cm]{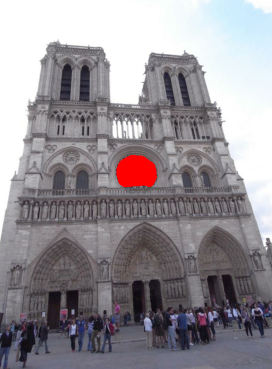}}{\textit{}}
    \jsubfig{\includegraphics[height=2.85cm]{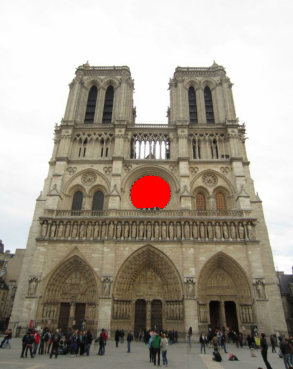}}{\textit{}}
    \jsubfig{\includegraphics[height=2.85cm]{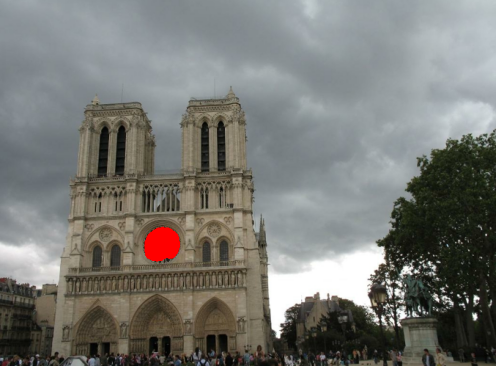}}{\textit{}}

\caption{\textbf{Results before and after 3D localization.} Segmentation results for the prompts \textit{windows} and \textit{rose window} are presented in the first and last pairs of rows, respectively. We show the results of \ourclipseg{} and \ourmethod{}'s projected localization in green, observing that \ourmethod{} yields 3D-consistent results by fusing the 2D predictions of \ourclipseg{}, which exhibit view inconsistencies. We also show binary segmentation (th) obtained with threshold $0.5$ in red, seeing that inconsistencies are prominent when using these methods for binary prediction.
}
\label{fig:consistency}
\end{figure*}

\subsection{Pseudo-Label Statistics} \label{sec:supp_pl_stats}

The pseudo-labels used in training consist of 4,031 unique non-empty values (over 58K images). The most common pseudo-labels are \emph{facade} (5,380 occurrences), \emph{dome} (3,084 occurrences), \emph{stained class windows} (2,550 occurrences), \emph{exterior} (2,365 occurrences), and \emph{interior} (1,649 occurences). 2,453 pseudo-labels occur only once (61\% of unique values) and 3,426 occur at most five times (79\% of unique values). Examples of pseudo-labels that only occur once include: \emph{spiral relief, the attic, elevation and vault, archevêché, goose tower, pentcost cross, transept and croisée}.

We note the long tail of pseudo-labels includes items shown in our evaluation such as \emph{tympanum} (29 occurrences), \emph{roundel} (occurs once as \emph{painted roundel}), \emph{colonnade} (230 occurrences), and \emph{pediment} (3 occurrences; 44 times as plural \emph{pediments}).

\subsection{CLIPSeg Visualizations}
\begin{figure}
\rotatebox{90}{\whitetxt{xxxx}Input}
  \jsubfig{\includegraphics[height=2.05cm]{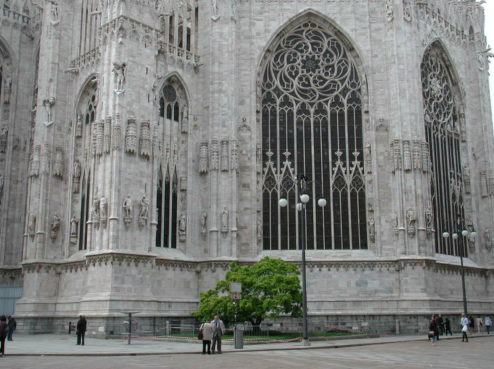}}{  }
  \hfill
   \jsubfig{\includegraphics[height=2.05cm]{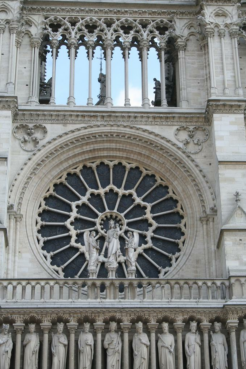}}{  }
   \hfill
    \jsubfig{\includegraphics[height=2.05cm]{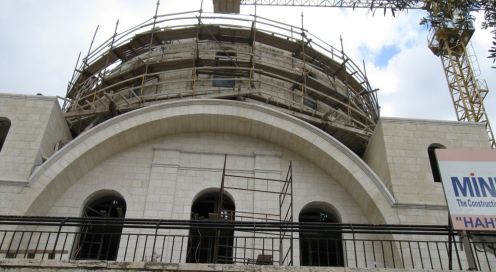}}{  }
 \\
\rotatebox{90}{\whitetxt{xxp}CLIPSeg}
  \jsubfig{\includegraphics[height=2.05cm]{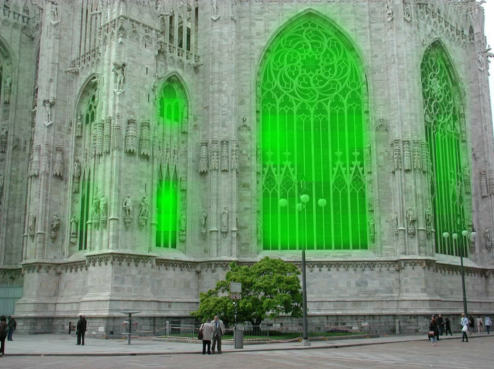}}{  }
  \hfill
   \jsubfig{\includegraphics[height=2.05cm]{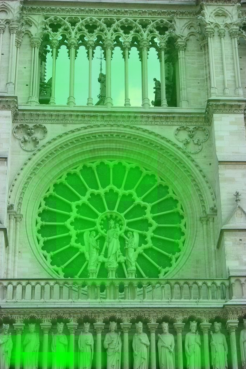}}{  }
   \hfill
    \jsubfig{\includegraphics[height=2.05cm]{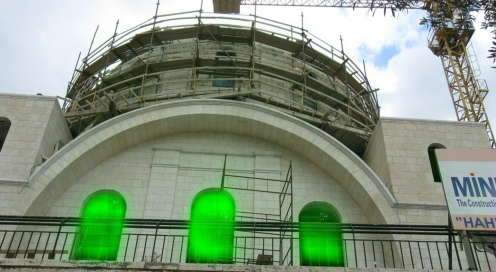}}{  }
 \\
 \rotatebox{90}{\whitetxt{xp}\ourclipseg{}}
  \jsubfig{\includegraphics[height=2.05cm]{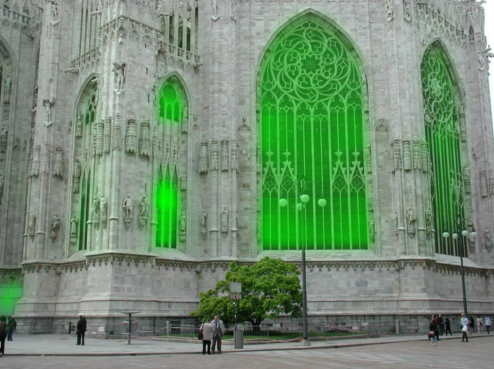}}{  }
  \hfill
   \jsubfig{\includegraphics[height=2.05cm]{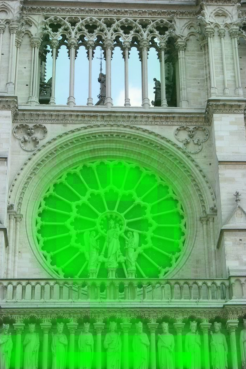}}{  }
   \hfill
    \jsubfig{\includegraphics[height=2.05cm]{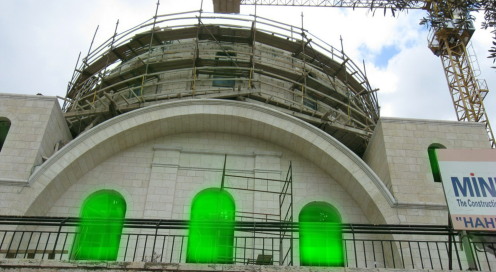}}{  }
    \caption{
  \textbf{Providing text-based segmentation models with partially related text prompts}. Above we provide the target prompt \textit{door} to CLIPSeg (pretrained and fine-tuned) along with images that do not have visible doors. As seen above, the models instead segment more salient regions which bear some visual and semantic similarity to the provided text prompt (in this case, segmenting windows).
  }\label{fig:adaptation_supp}
\end{figure}

As described in our main paper, we leverage the ability of CLIPSeg to segment salient objects in zoomed-in images even when it lacks fine-grained understanding of the accompanying pseudo-label. To illustrate this, Figure \ref{fig:adaptation_supp} shows several results of inputting the target text prompt \emph{door} to CLIPSeg along with images that do not have visible doors. As seen there, the model segments salient regions which bear some visual and semantic similarity to the provided text prompt (\emph{i.e.} possibly recognizing an ``opening'' agnostic to its fine-grained categorization as a door, portal, window, etc). Our fine-tuning scheme leverages this capability to bootstrap segmentation knowledge in zoomed-out views by supervising over zoomed-in views where the salient region is known to correspond to its textual pseudo-label.

Additionally, we find that 2D segmentation maps often show a bias towards objects and regions in the center of images, at the expense of the peripheries of scenes. This is seen for instance in Figure \ref{fig:adaptation_supp}, where the windows on the center are better localized, in comparison to the windows on the sides of the building. %

\subsection{\ourmethod{} Visualizations}

In Figure \ref{fig:consistency}, we compare segmentation results before and after 3D localization. We see that \ourmethod{} exhibits 3D consistency, while 2D segmentation results of \ourclipseg{} operating on each image separately exhibit inconsistent results between views. We also see that this effect is prominent when using these methods for binary segmentation, obtained by thresholding predictions.

\begin{figure}
\centering
    \jsubfig{\includegraphics[height=1.18cm]{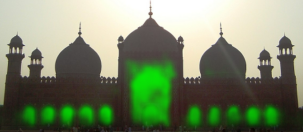}}{\textit{Portals}}
    \hfill
    \jsubfig{\includegraphics[height=1.18cm]{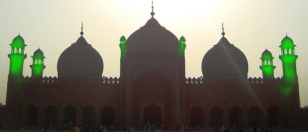}}{\textit{Minarets}}
    \hfill
    \jsubfig{\includegraphics[height=1.18cm]{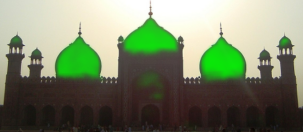}}{\textit{Domes}}
    \\
    \vspace{5pt}
    \jsubfig{\includegraphics[height=1.8cm]{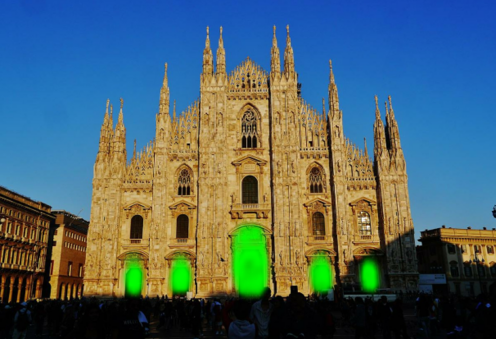}}{\textit{Portals}}
    \hfill
    \jsubfig{\includegraphics[height=1.8cm]{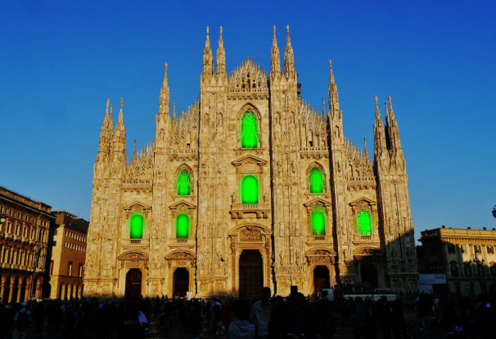}}{\textit{Windows}}
    \hfill
    \jsubfig{\includegraphics[height=1.8cm]{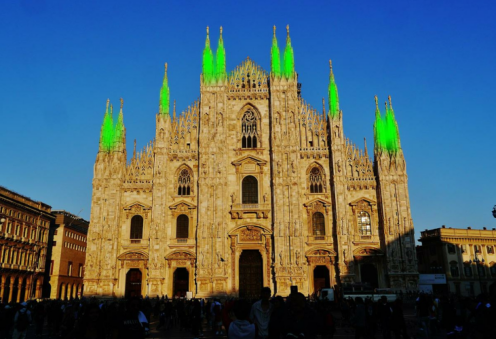}}{\textit{Spires}}
    
\caption{\textbf{Different localization images for the same scene.} We show multiple semantic concept localizations of \ourmethod{} for a single scene view. These results on Badshahi Mosque (first row) and Milan Cathedral (second row) illustrate how the user may provide \ourmethod{} with multiple text prompts to understand the semantic decomposition of a scene.
}
\label{fig:same_scene}
\end{figure}

\begin{figure}
\rotatebox{90}{\whitetxt{xxxxp}Input}
  \jsubfig{\includegraphics[height=2.1cm]{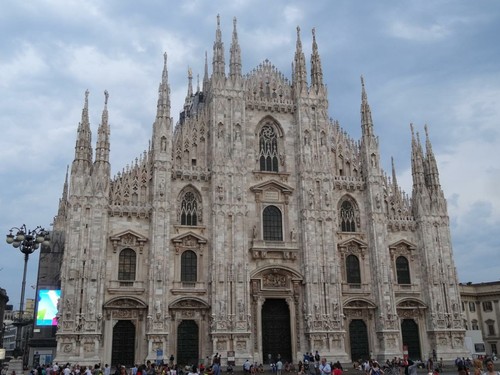}}{}
  \hfill
   \jsubfig{\includegraphics[height=2.1cm]{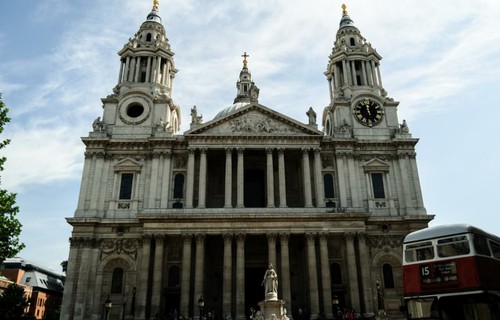}}{  }
   \hfill
    \jsubfig{\includegraphics[height=2.1cm]{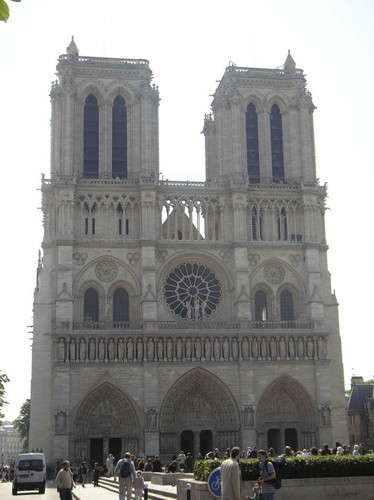}}{  }
 \\
\rotatebox{90}{\whitetxt{xxxxp}LSeg}
  \jsubfig{\includegraphics[height=2.1cm]{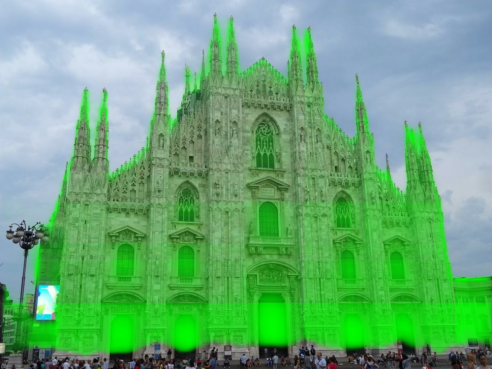}}{}
  \hfill
   \jsubfig{\includegraphics[height=2.1cm]{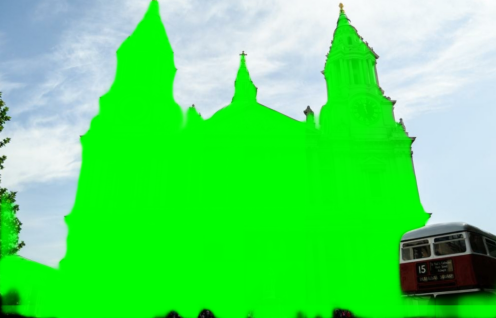}}{}
   \hfill
    \jsubfig{\includegraphics[height=2.1cm]{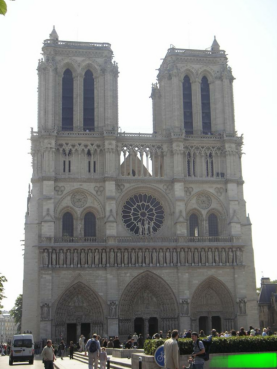}}{}
 \\
 \rotatebox{90}{\whitetxt{xxxpp}ToB{}}
  \jsubfig{\includegraphics[height=2.1cm]{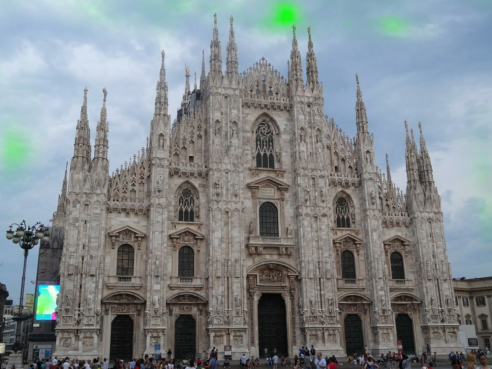}}{}
  \hfill
   \jsubfig{\includegraphics[height=2.1cm]{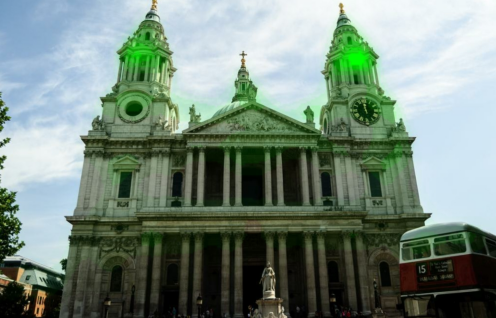}}{}
   \hfill
    \jsubfig{\includegraphics[height=2.1cm]{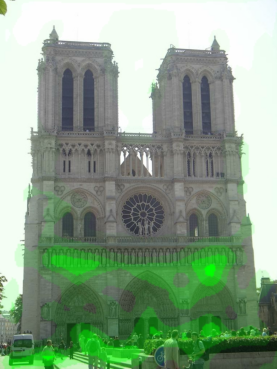}}{}
\rotatebox{90}{\whitetxt{xxp}CLIPSeg{}}
  \jsubfig{\includegraphics[height=2.1cm]{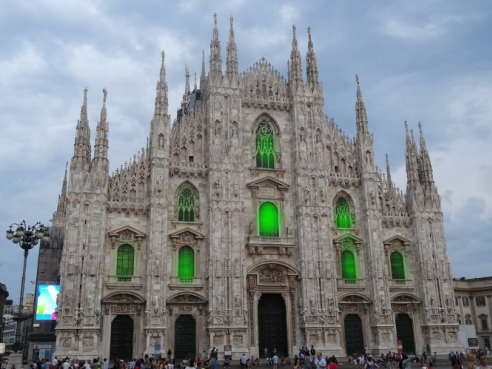}}{}
  \hfill
   \jsubfig{\includegraphics[height=2.1cm]{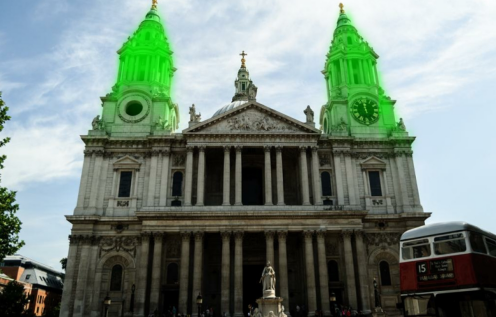}}{}
   \hfill
    \jsubfig{\includegraphics[height=2.1cm]{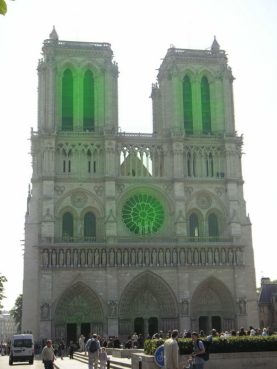}}{}
     \rotatebox{90}{\whitetxt{xx}\ourclipseg{}{}}
  \jsubfig{\includegraphics[height=2.1cm]{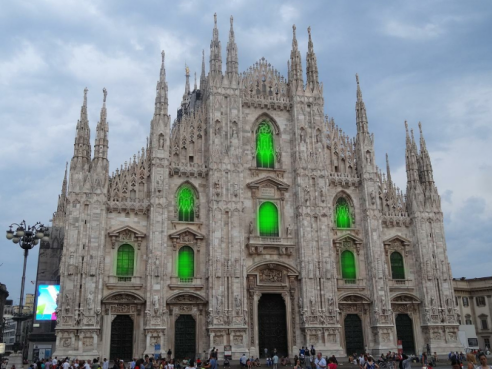}}{Windows}
  \hfill
   \jsubfig{\includegraphics[height=2.1cm]{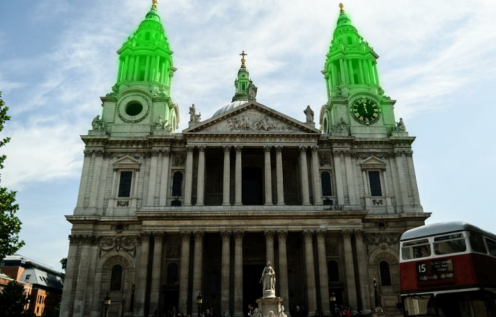}}{Towers}
   \hfill
    \jsubfig{\includegraphics[height=2.1cm]{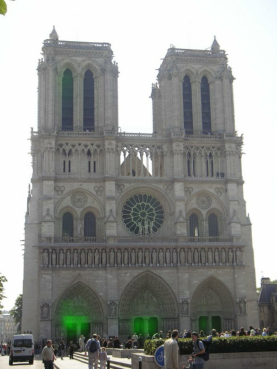}}{Portals}
    \caption{\textbf{Illustration of baseline 2D segmentation methods.} As is seen above, the baseline methods (LSeg and ToB) struggle to attend to the relevant regions in the images, while \ourclipseg{} shows the best understanding of these concepts and their localizations, consistent with our quantitative evaluation.
    }
    \label{fig:2d_baselines_supp}
\end{figure}

In Figure \ref{fig:same_scene}, we demonstrate our ability to perform localization of multiple semantic concepts in a single view of a scene. By providing \ourmethod{} with different text prompts, the user may decompose a scene into semantic regions to gain an understanding of its composition.

\subsection{2D Baseline Visualizations}

In Figure \ref{fig:2d_baselines_supp}, we visualize outputs of the two 2D baseline segmentation methods (LSeg and ToB) as well as CLIPSeg and our fine-tuned \ourclipseg{}. We see that the baseline methods struggle to attend to the relevant regions in our images, while \ourclipseg{} shows the best undestanding of these concepts and their localizations.

\subsection{\new{LLM Ablations}}

\new{As an additional test of our LLM-based pseudo-labeling procedure, we ablate the effect of the LLM model size and prompt templates used. In particular, we test the following sizes of Flan-T5~\cite{chung2022scaling}: XL (ours), Large, Base, and Small\footnote{Available on Hugging Face Model Hub at the following checkpoints: \texttt{google/flan-t5-xl}, \texttt{google/flan-t5-large}, \texttt{google/flan-t5-base}, \texttt{google/flan-t5-small}.} These vary in size from 80M (Small) to 3B (XL) parameters. In addition, we test the following prompt templates: $P_1$ (our original prompt, including the phrase \emph{...what architectural feature of...}), $P_2$ (\emph{...what aspect of the building...}), and $P_3$ (\emph{...what thing in...}).}

\new{We sample 100 random items from our dataset for manual inspection, running pseudo-labeling with our original setting (XL, $P_1$) as well as with alternate model sizes and prompts. Regarding model sizes, while the majority of non-empty generated pseudo-labels are valid as we show in the main paper, we consider how often empty or incorrect pseudo-labels are yielded when varying the model size. Considering this, 62/100 items receive an empty, poor or vague pseudo-label in our original setting, only one of these receives a valid pseudo-label with a smaller model, confirming the superior performance of the largest (XL) model. Regarding prompt variants, $P_2$ only yields 9/100 valid pseudo-labels (versus 38/100 for $P_1$), while $P_3$ yields 40/100 valid pseudo-labels (31 of these are in common with $P_1$). Thus, the best-performing prompt ($P_3$) is comparable to our original setting, suggesting that our original setting is well-designed to produce useful pseudo-labels.}

\subsection{\new{\ourclip{} Retrieval Results}}

\begin{table}[t]
  \centering
  \setlength{\tabcolsep}{3.8pt}
  \begin{tabularx}{0.42\textwidth}{@{ } llcccccc @{}}
    \toprule
    Method & {R@1} & {R@5} & {R@10} & {R@16} & {R@32} & {R@64} \\
    \midrule
    CLIP  & 0.04 & 0.18 & 0.24 & 0.31 & 0.47 & 0.68 \\
    \hl{\ourclip{}}  & \textbf{0.08} & \textbf{0.30} & \textbf{0.44} & \textbf{0.52} & \textbf{0.66} & \textbf{0.79} \\
    \bottomrule
  \end{tabularx}
  \caption{\new{\textbf{Terminology Retrieval Evaluation}. We evaluate image-to-text retrieval for finding relevant textual terms for test images. We report recall at $k\in\{1,5,10,16,32,64\}$, comparing our results (\hl{highlighted} in the table) to the baseline CLIP model. Best results are highlighted in \textbf{bold}.}}
\label{tab:retrieval}
\end{table}

\new{In Table \ref{tab:retrieval}, we show quantitative results for the use of \ourclip{} to retrieve relevant terminology, as described in our main paper. In particular, we fix a vocabulary of architectural terms found at least 10 times in the training data, and evaluate text-to-image retrieval on test images (from landmarks not seen during training) with pseudo-labels in this list. As seen in these results, our fine-tuning provides a significant performance boost to CLIP in retrieving relevant terms for scene views, as the base CLIP model is not necessarily familiar with fine-grained architectural terminology relevant to our landmarks out-of-the-box.}

\subsection{\new{Additional \ourclipseg{} Results}}

\begin{table}[t]
  \centering
  \setlength{\tabcolsep}{3.8pt}
  \begin{tabularx}{0.36\textwidth}{@{ } lcccc @{}}
    \toprule
    Method & Thresh. & IoU & Precision & Recall \\
    \midrule
    CLIPSeg & 0.10 & 0.634 & 0.807 & 0.813 \\
    \ourclipseg{} & 0.10 & \textbf{0.676} & \textbf{0.824} & \textbf{0.855} \\
    \midrule
    CLIPSeg & 0.15 & 0.650 & 0.815 & 0.827 \\
    \ourclipseg{} & 0.15 & \textbf{0.690} & \textbf{0.833} & \textbf{0.860} \\
    \midrule
    CLIPSeg & 0.20 & 0.653 & 0.818 & 0.832 \\
    \ourclipseg{} & 0.20 & \textbf{0.681} & \textbf{0.836} & \textbf{0.853} \\
    \midrule
    CLIPSeg & 0.25 & 0.649 & 0.817 & 0.832 \\
    \ourclipseg{} & 0.25 & \textbf{0.652} & \textbf{0.836} & \textbf{0.833} \\
    \bottomrule
  \end{tabularx}
  \caption{\new{Results on Wikiscenes~\cite{wu2021towers} for CLIPSeg before and after our fine-tuning procedure. Following Wu et al, we report IoU, precision, and recall scores. As these are threshold-dependent, we test multiple thresholds, finding that \ourclipseg{} shows a performance boost overall.}}
\label{tab:tob}
\end{table}

\new{To test the robustness of our CLIPSeg fine-tuning on additional datasets and preservation of pretraining knowledge, we evaluate segmentation results on two additional datasets: SceneParse150~\cite{zhou2016semantic,zhou2017scene} (general outdoor scene segmentation) and Wikiscenes~\cite{wu2021towers} (architectural terminology).}

\new{On SceneParse150, we test on the validation split (2000 items), selecting a random semantic class per image (from among those classes present in the image's annotations). We segment using the class's textual name and measure average precision, averaged over all items to yield the mean average precision (mAP) metric. We observe a negligible performance degradation after fine-tuning, namely mAP 0.53 before fine-tuning and 0.52 afterwards, suggesting overall preservation of pretraining knowledge.}

\new{On Wikiscenes, fine-tuning improves all metrics reported by Wu et al. (IoU, precision, recall), as shown in Table \ref{tab:tob}. As these metrics are threshold-dependent, we test multiple threshold values, and see that \ourclipseg{} shows an overall improvement in performance (e.g. reaching IoU of $0.890$ for the optimal threshold, while CLIPSeg without fine-tuning does not exceed $0.681$). Thus, as expected, our model specializes in the architectural domain while still showing knowledge of general terms from pretraining.}

\begin{table}[t]
  \centering
  \setlength{\tabcolsep}{3.2pt}
  \begin{tabularx}{0.99\columnwidth}{@{ } llccccccc @{}}
    \toprule
    Method & mAP & portal & window & spire & tower & dome & minaret \\
    \midrule
    LERF$^*$  & 0.14 & 0.16 & \textbf{0.15} & 0.18 & \textbf{0.13} & 0.10 & 0.09\\
    \ourlerf{}$^*$  & \textbf{0.15} & \textbf{0.18} & \textbf{0.15} & \textbf{0.19} & 0.12 & \textbf{0.12} & \textbf{0.11}\\
    \bottomrule
  \end{tabularx}
{\begin{flushleft}
  \footnotesize $^*$Using a Ha-NeRF backbone
\end{flushleft}}
\vspace{-2pt}
  \caption{\new{\textbf{LERF Comparison}. We report mean average precision (mAP; averaged per category) and per category average precision over the \ourdataset{} benchmark, comparing LERF with \ourlerf{}. Best results are highlighted in \textbf{bold}.}}
\label{tab:lerf}
\end{table}

\subsection{\new{LERF with \ourclip{}}}
\new{In Table \ref{tab:lerf}, we show quantitative results of LERF \cite{kerr2023lerf} using our \ourclip{}, and we compare it the results of LERF with CLIP without fine-tuning. We denote LERF with \ourclip{} as \ourlerf{}. In both cases, we use the Ha-NeRF backbone.
We see that the results of \ourlerf{} are only slightly better than the results of LERF with the original CLIP, suggesting that using better features for regression is not sufficient in our problem setting.}

\begin{table}[t]
  \centering
  \setlength{\tabcolsep}{3.2pt}
  \begin{tabularx}{0.67\columnwidth}{@{ } llccccccc @{}}
    \toprule
    Method & mAP & portal & window & spire \\
    \midrule
    DFF  & 0.07 & 0.04 & 0.05 & 0.12\\
    LERF & 0.19 & 0.32 & 0.06 & 0.2\\
     \ourmethod{}  & \textbf{0.65} & \textbf{0.76} & \textbf{0.56} & \textbf{0.64}\\
    \bottomrule
  \end{tabularx}
  \caption{\new{\textbf{Constant Illumination Comparison}. We report mean average precision (mAP; averaged per category) and per category average precision over the relevant categories in Milan Cathedral from Google-Earth using constant illumination. Best results are highlighted in \textbf{bold}.}}
\label{tab:const_illumination}
\end{table}

\begin{figure}
\rotatebox{90}{\whitetxt{xxp}DFF}
  \jsubfig{\includegraphics[height=1.45cm]{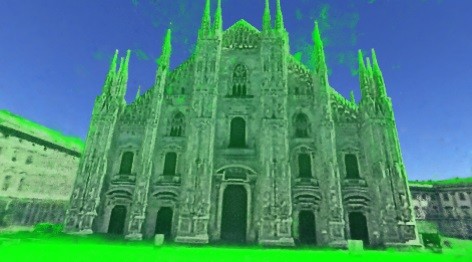}}{  }
  \hfill
   \jsubfig{\includegraphics[height=1.45cm]{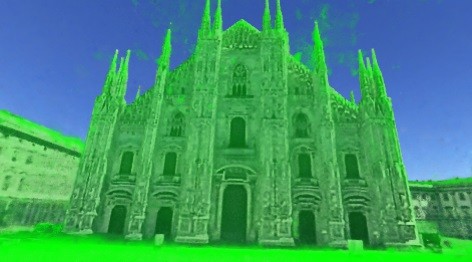}}{  }
   \hfill
    \jsubfig{\includegraphics[height=1.45cm]{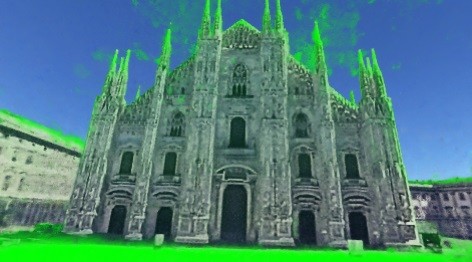}}{  }
       \hfill
 \\
    \rotatebox{90}{\whitetxt{xxp}LERF}
  \jsubfig{\includegraphics[height=1.45cm]{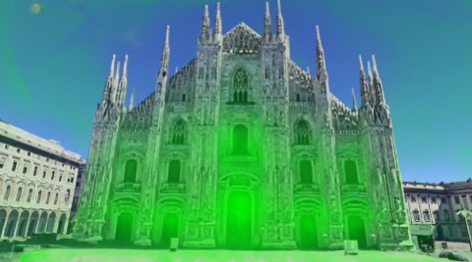}}{  }
  \hfill
   \jsubfig{\includegraphics[height=1.45cm]{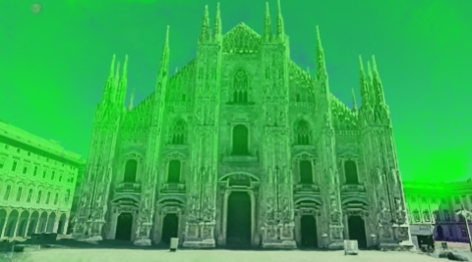}}{  }
   \hfill
    \jsubfig{\includegraphics[height=1.45cm]{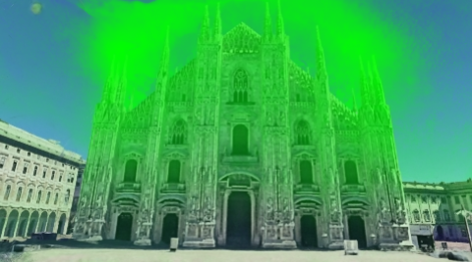}}{  }
       \hfill
 \\
 \rotatebox{90}{\whitetxt{xxp}Ours}
  \jsubfig{\includegraphics[height=1.45cm]{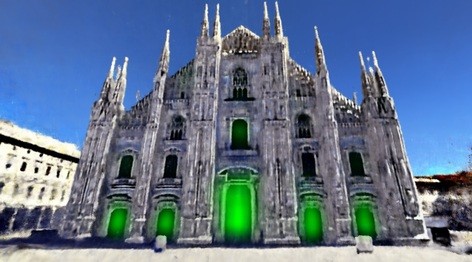}}{ \textit{Portal} }
  \hfill
   \jsubfig{\includegraphics[height=1.45cm]{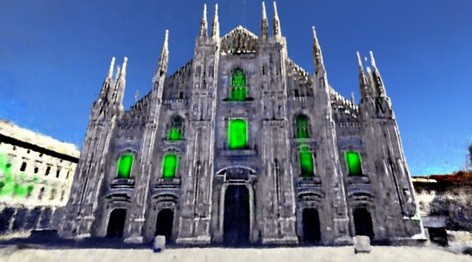}}{ \textit{Window} }
   \hfill
    \jsubfig{\includegraphics[height=1.45cm]{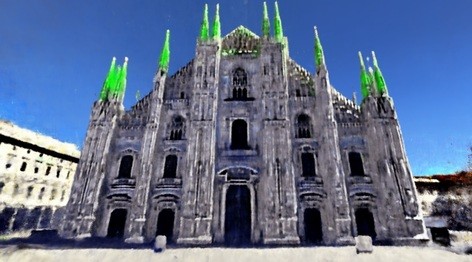}}{\textit{Spire}  }
   \hfill
  \caption{\new{
  \textbf{Localization comparison in a constant illumination setting}. Above we show localization for DFF, LERF, and \ourmethod{} (Ours) in a constant illumination setting, using input images rendered from Google Earth, as detailed further in the text. For this comparison, localization is performed using the original DFF and LERF implementations (and not our modified versions). As illustrated above, \ourmethod{} outperforms the other methods also in a constant illumination setting.}
  }\label{fig:constant_illumination}
\end{figure}

\subsection{\new{Results in a Constant Illumination Setting}}
\new{In Table \ref{tab:const_illumination}, we show quantitative results of DFF, LERF, and \ourmethod{} for a scene with constant illumination using a single camera. We used images rendered from Google Earth (following the procedure described in \cite{xiangli2021citynerf})
for the Milan Cathedral with the following three semantic categories: portal, window, and spire. Because the images were taken using a single camera with constant illumination, which adheres to the original DFF and LERF setting, we used their official public implementations.
We produced ground-truth binary segmentation maps to evaluate the results by manual labelling of five images per category. As the table shows, the results of \ourmethod{} are much better than those of DFF and LERF, even in the constant illumination case. These results further illustrate that these feature field regression methods are less effective for large-scale scenes. See Figure \ref{fig:constant_illumination} for a qualitative comparison in this setting.
}

\end{document}